\documentclass[lettersize,journal]{IEEEtran}
\usepackage{graphicx} 
\usepackage{amsmath}
\usepackage{amssymb}
\usepackage{booktabs}
\usepackage{color}
\usepackage{graphicx} 
\usepackage{pdfpages} 

\usepackage{enumerate}

\usepackage[ruled,vlined,linesnumbered]{algorithm2e}
\SetKwInput{KwIn}{Input}
\SetKwInput{KwOut}{Output}
\SetKw{Return}{return}
\SetKw{Break}{break}
\SetKw{Continue}{continue}

\usepackage{multirow}
\usepackage[table]{xcolor}

\usepackage{amssymb}
\usepackage{amsmath}
\usepackage{subfigure}
\usepackage{setspace}
\usepackage{float}

\title{KinoUAV: A Benchmark for Safe and Stable UAV Vision-Language Navigation with Kinodynamic Modeling}

\author{Chenyao Yang, Peijin Wang, Yingchao Feng, and Wenhui Diao%
	\thanks{This work was supported in part by the National Key R\&D Program of China under Grant 2024YFF1401001, in part by the National Natural Science Foundation of China under Grant 62301538, in part by the Science and Disruptive Technology Program under Grant 2025-AIRCAS-SDTP-04, and in part by the Youth Deployment Project of the Chinese Academy of Sciences under Grant GFQN-07. \textit{(Corresponding author: Peijin Wang.)}}%
	\thanks{Chenyao Yang and Peijin Wang are with the Aerospace Information Research Institute, Chinese Academy of Sciences, Beijing 100190, China; the University of Chinese Academy of Sciences, Beijing 100190, China; the School of Electronic, Electrical and Communication Engineering, University of Chinese Academy of Sciences, Beijing 100190, China; and the Key Laboratory of Target Cognition and Application Technology, Beijing 100190, China (e-mail: wangpeijin17@mails.ucas.ac.cn).}%
	\thanks{Yingchao Feng and Wenhui Diao are with the Aerospace Information Research Institute, Chinese Academy of Sciences, Beijing 100190, China, and the Key Laboratory of Target Cognition and Application Technology, Beijing 100190, China.}%
}

\begin{document}

\maketitle

\begin{abstract}
Unmanned aerial vehicles (UAVs) have recently emerged as important remote sensing platforms, and UAV navigation has shown great potential in remote sensing applications. However, existing methods do not explicitly incorporate kinodynamic modeling or safety distances into trajectory generation, resulting in high absolute yaw rates, frequent collisions, and low navigation success rates, which limit safety and motion stability. To address this challenge, we introduce KinoUAV, a kinodynamic benchmark comprising a dataset and KinoUAV-Agent. First, we develop a hierarchical kinodynamic trajectory generation pipeline that drives UAV through acceleration actions. It uses waypoint chains from global planning and generates trajectories segment by segment under velocity, acceleration, and jerk constraints, reducing oscillations and improving trajectory stability. Furthermore, we introduce directional free-space distance (DFD) labels into the constructed dataset to describe the available clearance for safe motion decisions. The proposed KinoUAV-Agent predicts DFDs and uses them to generate safe actions, decoupling perception from action decision-making. This design enables autonomous obstacle avoidance and improves navigation safety. The benchmark incorporates the proposed Cruise-Seek task, in which the UAV first enters the target region and then searches for the target, reducing the difficulty of target search and improving navigation success. Experiments on KinoUAV show that KinoUAV-Agent outperforms existing methods, achieving the highest success rate while maintaining lower absolute yaw rates and collision rates. These results demonstrate the effectiveness of kinodynamic modeling for safe and stable UAV navigation. The code will be released at https://github.com/CabbageYang15596/KinoUAV
\end{abstract}


\section{Introduction}

As important remote sensing platforms, UAVs are widely used in agricultural monitoring, infrastructure inspection, and disaster assessment \cite{shakhatreh2019unmanned, sun2024evaluation}. Through autonomous navigation, UAVs can freely approach and observe specific targets, thereby demonstrating immense potential in the field of remote sensing. To achieve autonomous navigation in complex environments, Vision-Language Navigation (VLN) has emerged as an important research direction in embodied intelligence. Early VLN studies mainly focused on ground-based agents \cite{anderson2018vision, ku2020room, krantz2020beyond, qi2020reverie, chen2019touchdown, thomason2020vision, shridhar2020alfred, fan2023aerial}. With the development of UAV simulation platforms, researchers have extended VLN to aerial environments \cite{liu2023aerialvln, wang2025towards, gao2025openfly, xiao2025uav}, greatly improving the scale of scenes and the amount of data. 

\begin{figure}[t]
	\centering
	\includegraphics[width=\columnwidth]{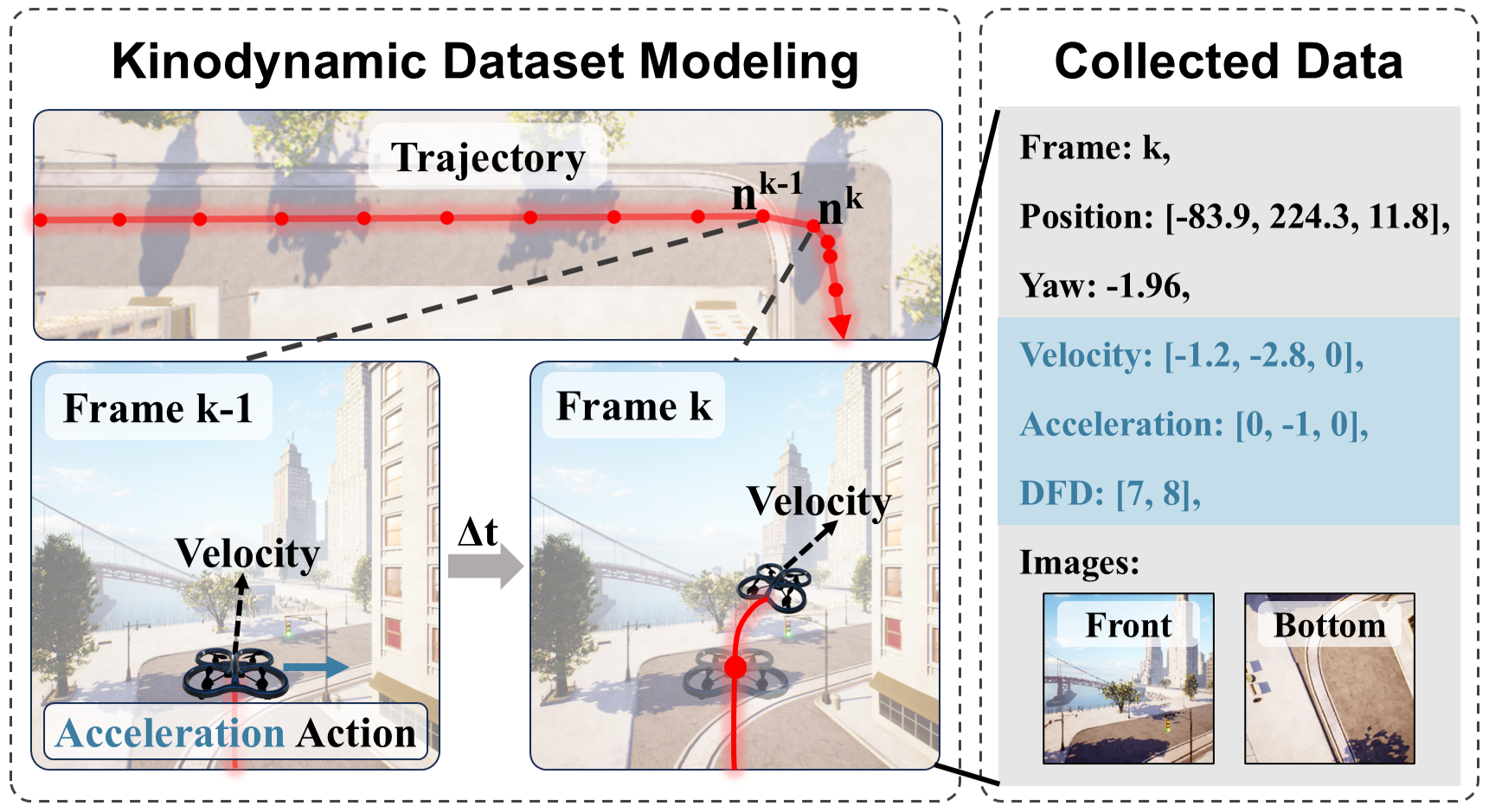}
	\caption{Motivation of KinoUAV. Kinodynamic modeling is introduced into the dataset construction pipeline. Acceleration-driven control is used to generate stable UAV trajectories. Motion states, DFDs, front and bottom-view images are collected at each frame. The blue region highlights the data uniquely provided by KinoUAV.}
	
	\label{fig1}
\end{figure}

However, existing UAV VLN benchmarks still suffer from several limitations in dataset construction. First, trajectories are commonly generated using discrete actions, waypoint interpolation, or loss-based motion constraints \cite{gao2025openfly, wang2025towards, jiang2026dynfly}, without explicitly integrating kinodynamic modeling into the generation. Such trajectories may exhibit abrupt motion changes and limited control precision, making them unsuitable for learning stable UAV motion. Second, existing datasets lack effective supervision for autonomous obstacle avoidance. Current datasets often introduce additional information, such as multi-source observations, known obstacle positions, or explicit instruction cues for collision avoidance \cite{wu2025aeroduo, li2025skyvln, gao2025openfly}. The substantial overhead of data collection and processing limits autonomous perception. Third, data containing detailed path instructions weaken autonomous exploration \cite{gao2025openfly}, while target descriptions alone lead to inefficient search in large-scale environments \cite{xiao2025uav}. Both types of data fail to provide guidance aligned with the current navigation stage, increasing the risk of missed targets and navigation failure.

To address these issues, we propose KinoUAV, a UAV VLN benchmark built with kinodynamic modeling. First, we design a trajectory generation pipeline based on hierarchical kinodynamic A*. The proposed method performs path search according to different waypoint chains and uses acceleration actions to control UAV motion. It generates stable flight trajectories segment by segment under velocity, acceleration and jerk constraints. We then collect images and motion states at each sampling point. To support safe navigation, we introduce forward and downward DFDs to quantify the obstacle-free ranges ahead of and below the UAV. These DFDs serve as intermediate supervision, enabling the model to perceive environmental risks before making action decisions. This design decouples environmental perception from action prediction.

Based on this data generation pipeline, we construct Main Subset and Auxiliary Subset: Auxiliary Subset uses randomly generated waypoint chains to plan trajectories, strengthening the UAV's waypoint tracking capability. The KinoUAV Dataset generates trajectories based on the start and target positions, enabling complete navigation capability. It is collected under the Cruise-Seek task setting to balance navigation guidance and autonomous exploration. During the Cruise stage, only target-region information is provided to guide the UAV for region-level navigation. After entering the target region, the task switches to the Seek stage, where target descriptions are provided for accurate target localization. This design avoids excessive path guidance and large-scale blind exploration.

Furthermore, we propose KinoUAV-Agent, which adopts a hierarchical navigation architecture. The high-level waypoint planner predicts the next waypoint during both the Cruise and Seek stages. The low-level motion controller uses estimated DFDs for autonomous obstacle avoidance and predicts acceleration actions for waypoint tracking.

Beyond existing VLN metrics, KinoUAV introduces absolute yaw rate to evaluate motion stability. Experimental results show that existing methods still face significant challenges in UAV VLN tasks with kinodynamic constraints. Meanwhile, KinoUAV-Agent achieves the highest success rate of 89.3\% while obtaining the lower collision rate and yaw rate, demonstrating the importance of kinodynamically modeled data and comprehensive evaluation protocols for advancing safe and stable UAV VLN research. KinoUAV provides a unified dataset and evaluation standard for future research. Our main contributions are summarized as follows:

\begin{itemize}
\item KinoUAV is among the first UAV VLN benchmarks to introduce a kinodynamic pipeline for dataset construction, enabling acceleration-driven trajectory generation.
\item DFDs provide intermediate supervision for obstacle avoidance using only RGB images, decoupling perception from action prediction.
\item The Cruise-Seek task combines region-level navigation with target-level search, balancing navigation guidance and autonomous exploration.

\end{itemize}

\section{Related Work}

\subsection{Kinodynamic Motion Planning}
Motion planning aims to generate collision-free paths from map information. A* \cite{hart1968formal} provides a classical heuristic search framework that combines the accumulated path cost with an estimate of the remaining cost. However, geometric connectivity alone does not guarantee that a motion can be executed. Hybrid A* \cite{dolgov2008practical} retains continuous states within discrete search nodes and uses orientation-dependent state transitions to satisfy nonholonomic kinematic constraints. Beyond kinematic feasibility, Donald et al. \cite{donald1993kinodynamic} formulated kinodynamic planning under motion constraints, including limits on velocity and acceleration. State lattice planning \cite{pivtoraiko2009differentially} constructs a graph of feasible motion primitives to maintain state continuity and accommodate the vehicle's motion model. For aerial robots, Zhou et al. \cite{zhou2019robust} combined kinodynamic search with B-spline optimization and time adjustment to improve trajectory smoothness, obstacle avoidance, and dynamic feasibility. iDb-A* \cite{ortiz2024idb} alternates between motion-primitive search and trajectory optimization, allowing bounded discontinuities during search and repairing them during optimization. Following this principle, iDb-RRT \cite{ortiz2024idbrrt} combines sampling-based exploration with motion primitives and trajectory optimization. These methods provide a foundation for generating executable trajectories.

\subsection{VLN Datasets}
VLN benchmarks cover ground and aerial navigation, with tasks ranging from path following and object-oriented exploration to interactive navigation. Room-to-Room (R2R) \cite{anderson2018vision} established a benchmark for indoor VLN using human instructions and navigation trajectories in the Matterport3D simulation environment. Building on R2R, Room-for-Room (R4R) \cite{jain2019stay} concatenates paths to form longer routes and introduces instruction fidelity, while Room-Across-Room (RxR) \cite{ku2020room} expands multilingual coverage and provides dense spatiotemporal grounding between instructions and demonstrations. REVERIE \cite{qi2020reverie} moves beyond detailed route instructions by requiring agents to navigate to and identify a remotely described target object. SOON \cite{zhu2021soon} introduces scenario-oriented object search from arbitrary starting positions. To support interactive navigation, Cooperative Vision-and-Dialog Navigation (CVDN) \cite{thomason2020vision} collects human dialogues in which navigators request assistance from an oracle. Outdoor ground navigation is explored by LANI \cite{misra2018mapping}, which contains landmark-based instructions in synthetic environments, and Touchdown \cite{chen2019touchdown}, which combines street-level navigation with spatial grounding in real urban imagery. VLN-CE \cite{krantz2020beyond} transfers instruction following to continuous environments with low-level action control, addressing the simplifying assumptions of navigation graphs. Related embodied benchmarks, such as ALFRED \cite{shridhar2020alfred}, further combine navigation with object interaction and household task execution. These datasets offer diverse settings for ground VLN, but aerial navigation also requires reasoning about altitude changes and more complex three-dimensional spaces. AVDN \cite{fan2023aerial} extends dialogue-guided navigation to an aerial simulator built on xView and collects asynchronous dialogues between commanders and followers. AerialVLN \cite{liu2023aerialvln} provides an AirSim- and UE4-based platform for path following in three-dimensional aerial environments, using manually collected trajectories and route instructions. CityNav \cite{lee2025citynav} improves environmental realism with real-world point clouds from Cambridge and Birmingham and uses human flight trajectories to construct high-altitude UAV navigation tasks grounded in real-world data. TravelUAV \cite{wang2025towards} uses the OpenUAV platform to collect and annotate human flight trajectories, supporting navigation with target-direction information and assistance at different guidance levels. To reduce data collection costs, OpenFly \cite{gao2025openfly} integrates point-cloud acquisition, semantic annotation, A*-based trajectory generation, and automatic instruction generation into a pipeline for producing navigation data at scale. Together, these datasets have advanced VLN through varied task settings, simulated or real-world scenes, manually or automatically constructed trajectories, and diverse language annotations. Beyond these datasets, KinoUAV incorporates kinodynamic modeling into UAV navigation trajectory generation and records motion states to support learning safe and stable aerial navigation.

\subsection{VLN Methods}
VLN methods have progressed from recurrent instruction-following policies to pretrained multimodal models, structured spatial representations, and large-model-based navigation. Early ground-based approaches mainly improve language grounding and generalization. Speaker-Follower \cite{fried2018speaker} uses a speaker model to generate additional instructions and evaluate candidate routes, while EnvDrop \cite{tan2019learning} combines back translation with environmental dropout to improve transfer to unseen environments. Reinforced Cross-Modal Matching (RCM) \cite{wang2019reinforced} introduces local and global instruction-trajectory matching with self-supervised imitation learning. To track execution progress, the Self-Monitoring agent \cite{ma2019self} estimates instruction completion, and the Regretful Agent \cite{ma2019regretful} uses progress estimation to support backtracking and action revision. Subsequent studies strengthen cross-modal representations through pretraining. PREVALENT \cite{hao2020towards} learns representations from image-text-action triplets, whereas VLN-BERT \cite{majumdar2020improving} transfers visual grounding knowledge from web image-text pairs to instruction-path matching. Recurrent VLN-BERT \cite{hong2021vln} introduces a recurrent state into a multimodal Transformer for sequential decisions, and HAMT \cite{chen2021history} explicitly encodes long-horizon observations for history-aware action prediction. Beyond unstructured temporal memory, map-based methods model the spatial relationships needed for exploration. DUET \cite{chen2022think} combines global topological planning with local cross-modal grounding through a dual-scale graph Transformer. Cross-Modal Map Learning \cite{georgakis2022cross} predicts language-conditioned semantic maps and waypoint paths, while BEVBert \cite{an2022bevbert} integrates metric and topological maps through multimodal map pretraining. GridMM \cite{wang2023gridmm} organizes historical observations into a growing egocentric grid memory, and ETPNav \cite{an2024etpnav} combines evolving topological maps, cross-modal planning, and obstacle-aware execution in continuous environments. With the development of large language and vision-language models, recent methods increasingly exploit explicit reasoning and general visual knowledge. NavGPT \cite{zhou2024navgpt} performs zero-shot navigation through language-based reasoning, and NavGPT-2 \cite{zhou2024navgpt2} integrates large-model reasoning with a learned navigation policy. MapGPT \cite{chen2024mapgpt} introduces map-guided prompting and adaptive path planning, whereas InstructNav \cite{long2024instructnav} combines dynamic instruction planning with multiple value maps for diverse navigation tasks. Alongside these modular systems, NaVid \cite{zhang2024navid} uses a video-based VLM to predict navigation actions from visual history and instructions, and Uni-NaVid \cite{zhang2024uni} extends this formulation to multiple embodied navigation tasks. These ground-based studies provide useful mechanisms for semantic alignment, memory, exploration, and action prediction, which aerial methods adapt to larger spaces and additional motion dimensions. TravelUAV \cite{wang2025towards} employs a hierarchical decoder to predict a target pose and a refined local trajectory, connecting multimodal reasoning with local motion generation. FlightGPT \cite{cai2025flightgpt} combines supervised fine-tuning, reinforcement learning, and chain-of-thought reasoning to improve navigation decisions. GeoNav \cite{xu2026geonav} integrates a global cognitive map and a local hierarchical scene graph within landmark navigation, target search, and precise localization. AeroDuo \cite{wu2025aeroduo} exploits complementary high- and low-altitude observations, coordinating target reasoning with local navigation and grounding. FineCog-Nav \cite{shao2026finecog} organizes zero-shot navigation into cognitive modules with specialized prompts and structured interfaces. In parallel with advances in semantic reasoning, several studies address aerial motion execution. SkyVLN \cite{li2025skyvln} combines spatial verbalization and path memory with nonlinear model predictive control for trajectory tracking and dynamic obstacle avoidance. DynFly \cite{jiang2026dynfly} generates B-spline trajectories through flow matching and introduces motion-state supervision to improve trajectory continuity. AerialVLA \cite{xu2026aerialvla} combines front and downward observations with coarse directional instructions, predicts spatial offsets and yaw changes, and converts them into velocity commands. These developments show growing attention to both semantic navigation and executable motion. Our work connects these aspects through kinodynamically generated trajectories and acceleration-driven low-level control.

\begin{figure*}[t]
	\centering
	\includegraphics[width=\textwidth]{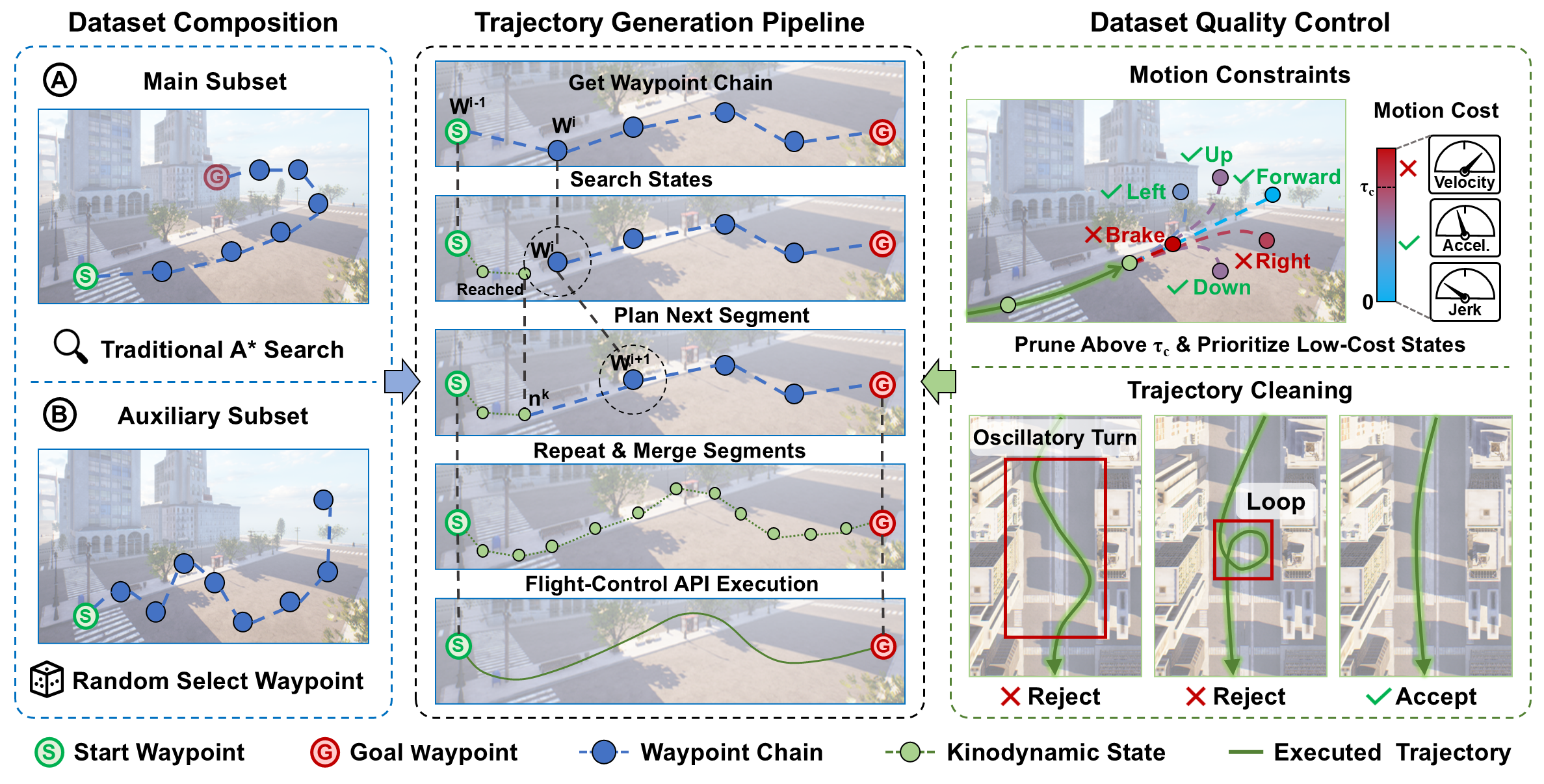}
	\caption{Overview of our dataset generation pipeline. Different waypoint chains are processed by hierarchical kinodynamic A$^*$ to generate reference trajectories under motion constraints. After trajectory cleanup, the reference trajectories are tracked through AirSim's flight-control API to collect observations and motion states for the Main Subset and Auxiliary Subset.}
	\label{fig:pipeline}
\end{figure*}

\section{Methods}

\subsection{Trajectory Generation Pipeline}

This pipeline consists of three stages. Searching directly over the kinodynamic state space for a long-range trajectory is computationally expensive. First, a waypoint chain is generated without kinematic constraints. Next, we use hierarchical kinematic A* algorithm to produce a sequence of discrete reference trajectory points based on the waypoint chain. Finally, the AirSim flight control API is invoked to generate a continuous trajectory. 

\subsubsection{Waypoint Chain Generation}

We acquire LiDAR point clouds for each scene and convert them into voxel occupancy maps \cite{gao2025openfly}. Each voxel is classified as occupied or free, and its side length is set to that of the smallest cube enclosing the UAV.

For the Main Subset, we specify a start position $\mathbf{p}_{\mathrm{s}}$ and a goal position $\mathbf{p}_{\mathrm{g}}$, and use geometric A$^*$ to search for a backbone path between them. At each expanded node, candidate positions are generated using a fixed step length $s_{\mathrm{step}}=3\,\mathrm{m}$. Let \(\Phi=\{0,\pm\frac{\pi}{6},\pm\frac{\pi}{3},\pm\frac{\pi}{2},\pm\frac{2\pi}{3},\pm\frac{5\pi}{6},\pi\}\) be the azimuth set and \(\Theta=\{\frac{\pi}{6},\frac{\pi}{3},\frac{\pi}{2},\frac{2\pi}{3},\frac{5\pi}{6}\}\) be the polar-angle set, where the polar angle is measured from the positive vertical axis. A candidate displacement is
\begin{equation}
	\Delta\mathbf{p}(\phi,\theta)=s_{\mathrm{step}}\begin{bmatrix}\sin\theta\cos\phi\\ \sin\theta\sin\phi\\ \cos\theta\end{bmatrix},\qquad \phi\in\Phi,\ \theta\in\Theta.
	\label{eq:coarse_motion}
\end{equation}
This sampling produces $60$ candidate directions per expanded node. Geometric A$^*$ searches over valid candidates and returns an ordered waypoint chain $\mathcal{W}=(\mathbf{q}_0,\mathbf{q}_1,\ldots,\mathbf{q}_M)$, where $\mathbf{q}_0=\mathbf{p}_{\mathrm{s}}$ and the final waypoint corresponds to the goal.

For Auxiliary Subset, we first sample a starting waypoint $\mathbf{q}_0$ in free space. Each subsequent waypoint is sampled relative to the preceding waypoint:
\begin{equation}
	\mathbf{q}_{i+1}=\mathbf{q}_i+r_i\begin{bmatrix}\sin\theta_i\cos\phi_i\\ \sin\theta_i\sin\phi_i\\ \cos\theta_i\end{bmatrix},\qquad 3<r_i<20\,\mathrm{m}.
	\label{eq:random_waypoints}
\end{equation}
Here, $\phi_i$ and $\theta_i$ are randomly sampled without the angular discretization. If the sampled waypoint lies in an occupied voxel, it is rejected and sampled again. This procedure continues until the chain contains ten waypoints, $\mathcal{W}=(\mathbf{q}_0,\ldots,\mathbf{q}_9)$, including the starting waypoint. For both subsets, the resulting waypoint chain defines the order of local targets for hierarchical kinodynamic A$^*$.

The Main Subset and Auxiliary Subset use different methods to generate their waypoint chains but share the same subsequent kinodynamic planner.

\subsubsection{Hierarchical Kinodynamic A$^*$ planning}

To generate UAV trajectories that account for velocity-dependent drag and changes in motion, we adopt a simplified kinodynamic model. Let \(m_{\mathrm{UAV}}\) be the UAV mass, and let \(\mathbf{p}^k\), \(\mathbf{v}^k\), \(\mathbf{a}^k\) and \(\mathbf{j}^k\) denote its position, velocity, total acceleration and jerk at step \(k\) in an inertial coordinate frame. Taking \(\mathbf{e}_z=[0,0,1]^\top\) as the upward unit vector and neglecting external disturbances, Newton's second law gives
\begin{equation}
	m_{\mathrm{UAV}}\mathbf{a}^k
	=\mathbf{F}_p^k+\mathbf{F}_d^k
	-m_{\mathrm{UAV}}g\mathbf{e}_z,
\end{equation}
where \(\mathbf{F}_p^k\) is the propulsion force and \(\mathbf{F}_d^k\) is the drag force. We assume that a low-level controller compensates for gravity and express the effective propulsion force as
\begin{equation}
	\mathbf{F}_p^k
	=m_{\mathrm{UAV}}g\mathbf{e}_z+\mathbf{F}_u^k,
\end{equation}
where \(\mathbf{F}_u^k\) is the gravity-compensated control force. We approximate aerodynamic drag using linear and quadratic velocity-dependent terms:
\begin{equation}
	\mathbf{F}_d^k
	=-m_{\mathrm{UAV}}
	\left(k_1\mathbf{v}^k
	+k_2\|\mathbf{v}^k\|\mathbf{v}^k\right),
\end{equation}
where \(k_1,k_2>0\) are mass-normalized drag coefficients. Substituting the propulsion and drag forces into the force balance yields
\begin{equation}
	\mathbf{a}^k=\frac{\mathbf{F}_u^k}{m_{\mathrm{UAV}}}-k_1\mathbf{v}^k-k_2\|\mathbf{v}^k\|\mathbf{v}^k.
\end{equation}
We write this as $\mathbf{a}^k=\mathbf{a}_{\mathrm{UAV}}^k+\mathbf{a}_d^k$, where $\mathbf{a}_{\mathrm{UAV}}^k=\mathbf{F}_u^k/m_{\mathrm{UAV}}$ is the control acceleration selected during planning, and $\mathbf{a}_d^k=-k_1\mathbf{v}^k-k_2\|\mathbf{v}^k\|\mathbf{v}^k$ is the drag acceleration determined by the current velocity.

During planning, the above model is used to propagate reference motion states at a fixed interval $\Delta t$. We use the subscript $\mathrm{ref}$ to distinguish these planned states from the realized UAV states obtained during flight. The reference velocity and position are updated as
\begin{equation}
	\begin{aligned}
		\mathbf{v}_{\mathrm{ref}}^{k+1}
		&=
		\mathbf{v}_{\mathrm{ref}}^{k}
		+\mathbf{a}_{\mathrm{ref}}^{k}\Delta t,\\
		\mathbf{p}_{\mathrm{ref}}^{k+1}
		&=
		\mathbf{p}_{\mathrm{ref}}^{k}
		+\mathbf{v}_{\mathrm{ref}}^{k+1}\Delta t,
	\end{aligned}
	\label{eq:state_propagation}
\end{equation}

where $\mathbf{a}_{\mathrm{ref}}^{k}$ is the total inertial frame acceleration computed by the simplified model. Each reference point specifies a desired position. The jerk used during planning is computed as
\begin{equation}
	\mathbf{j}_{\mathrm{ref}}^{k}
	=
	\frac{
		\mathbf{a}_{\mathrm{ref}}^{k}
		-
		\mathbf{a}_{\mathrm{ref}}^{k-1}
	}{\Delta t}.
	\label{eq:planning_jerk}
\end{equation}

For each target $\mathbf{q}_i$, $i=1,\ldots,M$, the local planner searches over states containing position, velocity, previous acceleration, and heading. For planning, we assume that this force can act along any of the six signed body-axis directions $\mathcal{E}=\{\pm\mathbf{e}_x,\pm\mathbf{e}_y,\pm\mathbf{e}_z\}$, with the same maximum magnitude $F_{\mathrm{UAV,max}}$, and let $\Gamma=\{0,0.5,1.0\}$ be the set of dimensionless force levels, in the body frame at step $k$. We define the legal control-acceleration set as
\begin{equation}
	\mathcal{A}_{\mathrm{UAV}}
	=
	\left\{
	\gamma\frac{F_{\mathrm{UAV,max}}}{m_{\mathrm{UAV}}}\mathbf{e}
	\;\middle|\;
	\gamma\in\Gamma,\ \mathbf{e}\in\mathcal{E}
	\right\}
\end{equation}

$\mathbf{a}_{\mathrm{UAV,ref}}^{k}\in\mathcal{A}_{\mathrm{UAV}}$. It results in 13 distinct candidates. Candidate states are rejected if they violate the speed or jerk limit or the applicable collision check. 

We evaluate each transition from $n^k$ to $n^{k+1}$ using the
kinodynamic cost
\begin{equation}
	\mathrm{cost}_{\mathrm{kin}}(n^k,n^{k+1})
	=\mathrm{cost}_{v}^{k}
	+\mathrm{cost}_{a}^{k}
	+\mathrm{cost}_{j}^{k},
	\label{eq:kin_cost}
\end{equation}
where
\begin{equation}
	\begin{aligned}
		\mathrm{cost}_{v}^{k}
		&= \lambda_1\|\mathbf{v}_{\mathrm{ref}}^{k}\| + \lambda_2 \angle\left(\mathbf{v}_{\mathrm{ref}}^{k},\ \mathbf{v}_{\mathrm{ref}}^{k+1}\right), \\
		\mathrm{cost}_{a}^{k}
		&= \lambda_3 f(\mathbf{a}_{\mathrm{UAV,ref}}^{k}), \\
		\mathrm{cost}_{j}^{k}
		&= \lambda_4\|\mathbf{j}_{\mathrm{ref}}^{k}\|.
	\end{aligned}
	\label{eq:kin_cost_terms}
\end{equation}

Here, $\angle\left(\mathbf{v}_{\mathrm{ref}}^{k},\ \mathbf{v}_{\mathrm{ref}}^{k+1}\right)$ is the angle between $\mathbf{v}_{\mathrm{ref}}^{k}$ and $\mathbf{v}_{\mathrm{ref}}^{k+1}$. The angle is set to zero if either velocity vector is zero. The second term penalizes selected special actions:
\begin{equation}
	f(\mathbf{u})=
	\begin{cases}
		1, & \mathbf{u}\in\mathcal{A}_{\mathrm{sp}},\\
		0,      & \text{otherwise},
	\end{cases}
	\quad
	\mathcal{A}_{\mathrm{sp}}
	=\{(0,1,0)^\top,(0,-1,0)^\top,\mathbf{0}\},
	\label{eq:special_action}
\end{equation}
The coefficients $\lambda_1,\ldots,\lambda_4$ control the relative contributions of these terms. Candidate nodes with $\mathrm{cost}_{\mathrm{kin}}>\tau_c$ are pruned, and the remaining nodes are ordered by their accumulated transition cost plus a distance-to-waypoint heuristic.

\begin{algorithm}[t]
	\caption{Hierarchical Kinodynamic A$^*$}
	\label{alg:hierarchical_astar}
	\footnotesize
	\KwIn{Start $\mathbf{p}_{\mathrm{s}}$, goal $\mathbf{p}_{\mathrm{g}}$, map $\mathcal{O}$, action set $\mathcal{A}$, cost threshold $\tau_c$}
	\KwOut{Trajectory $\mathcal{T}$ or failure}
	
	Generate waypoints $\{\mathbf{q}_1,\ldots,\mathbf{q}_M\}$ from $\mathbf{p}_{\mathrm{s}}$ to $\mathbf{p}_{\mathrm{g}}$ using geometric A$^*$\;
	\If{no path is found}{
		\Return failure\;
	}
	$\mathcal{T}\gets\varnothing$\;
	Initialize start state $n_0$ at $\mathbf{p}_{\mathrm{s}}$\;
	\For{$i\gets1$ \KwTo $M$}{
		Open set $Q\gets\{n_0\}$; closed set $C\gets\varnothing$\;
		$g(n_0)\gets0$; clear the parent of $n_0$\;
		$\mathrm{found}\gets\mathrm{false}$\;
		\While{$Q\neq\varnothing$}{
			Select $n$ with the smallest $g(n)+h(n)$ from $Q$, where $h(n)=\|\mathbf{p}(n)-\mathbf{q}_i\|$\;
			Remove $n$ from $Q$\;
			\If{the discretized state of $n$ is already in $C$}{
				\Continue\;
			}
			\If{$n$ satisfies the arrival condition at $\mathbf{q}_i$}{
				Trace parent states from $n$ back to $n_0$\;
				Append this segment to $\mathcal{T}$ in forward order, without repeating the connecting state\;
				$n_0\gets n$; $\mathrm{found}\gets\mathrm{true}$\;
				\Break\;
			}
			Add the discretized state of $n$ to $C$\;
			\ForEach{$\mathbf{u}\in\mathcal{A}$}{
				Rotate $\mathbf{u}$ from the body frame to the world frame\;
				Compute the next state $n'$ using the UAV dynamics\;
				\If{the motion collides with an obstacle or violates motion constraints}{
					\Continue\;
				}
				\If{the discretized state of $n'$ is in $C$}{
					\Continue\;
				}
				$c\gets\mathrm{cost}_{\mathrm{kin}}(n,n')$\;
				\If{$c>\tau_c$}{
					\Continue\;
				}
				$g_{\mathrm{new}}\gets g(n)+c$\;
				\If{$n'$ represents a new state or $g_{\mathrm{new}}$ is lower than its recorded cost}{
					$g(n')\gets g_{\mathrm{new}}$\;
					Set $n$ as the parent of $n'$\;
					Add or update $n'$ in $Q$\;
				}
			}
		}
		\If{$\mathrm{found}=\mathrm{false}$}{
			\Return failure\;
		}
	}
	\Return $\mathcal{T}$\;
\end{algorithm}

When the state enters the arrival radius $r_{\mathrm{arr}}$ of the currently active waypoint, the local search for that segment concludes, and the node information is recorded. The successful search segment is reconstructed by backtracking via parent pointers and appended to the completed trajectory segments. The terminal position, velocity and acceleration of this branch serve as the initial conditions for the search toward the next waypoint; this process continues until all waypoints have been reached. Finally, a trajectory cleanup step is performed to eliminate anomalous patterns such as oscillatory steering and loops.

\subsection{Dataset Construction}
We construct the dataset using the UE4 environment assets provided by OpenUAV \cite{wang2025towards}. We select 16 representative scenes and integrate hierarchical kinodynamic A$^*$ into the automated data collection pipeline \cite{xu2026aerialvla}. 

After generating the reference trajectory using hierarchical kinodynamic A$^*$, the trajectory $\mathrm{traj}=\{n^0,\ldots,n^{K-1}\}$ is then executed progressively through flight-control API. At each control step $k$, we compute a velocity command from the current UAV position $\mathbf{p}^k$ to the next reference position $\mathbf{p}_{\mathrm{ref}}^{k+1}$ using the time interval $\Delta t$:
\begin{equation}
	\mathbf{v}_{\mathrm{cmd}}^k
	=
	\frac{\mathbf{p}_{\mathrm{ref}}^{k+1}-\mathbf{p}^k}{\Delta t}.
\end{equation}
This velocity command is sent to the flight controller to guide the UAV toward the next reference point. The pose of the drone is obtained as:
\begin{equation}
	\boldsymbol{\eta}^{k}
	=
	\left[
	p_x^{k},\,p_y^{k},\,p_z^{k},\,
	\mathrm{pitch}^{k},\,\mathrm{roll}^{k},\,\mathrm{yaw}^{k}
	\right]^\top.
\end{equation}

At each valid control step, we record the realized UAV pose $\boldsymbol{\eta}^k$, velocity $\mathbf{v}^k$, associated reference control acceleration $\mathbf{a}_{\mathrm{UAV,ref}}^k$, and the rendered front- and bottom-view images $I_f^k$ and $I_b^k$. We also compute the corresponding front- and bottom-view directional free-space distances (DFDs), $d_f^k$ and $d_b^k$, from the occupancy map.

A DFD measures the distance from the current UAV position to the first occupied voxel intersected by a ray in the directions observed by the camera. Each ray has a maximum length of $100\,\mathrm{m}$. Ray traversal stops at the first occupied voxel, and the traveled distance is recorded as the DFD. If no occupied voxel is encountered within $100\,\mathrm{m}$, the DFD is assigned the maximum value of $100\,\mathrm{m}$.

The resulting data support the two levels of the navigation model.
At each high-level planning event, we form a waypoint-planning
sample
\begin{equation}
	D_h^i=
	\left(T^i,\mathbf{w}^i,I_f^{k_i},I_b^{k_i}\right),
\end{equation}
where $k_i$ is the control step at which body-frame waypoint $\mathbf{w}^i$ is planned. $T^i$ is the stage-dependent navigation instruction. For low-level control, each valid control step provides a sample
\begin{equation}
	D_l^k=
	\left(
	\boldsymbol{\eta}^k,\mathbf{v}^k,
	\mathbf{a}_{\mathrm{UAV,ref}}^k,
	\mathbf{w}^i,
	I_f^k,I_b^k,d_f^k,d_b^k
	\right).
\end{equation}
Thus, high-level samples pair observations and instructions with the next waypoint, whereas low-level samples pair the UAV state and observations with an acceleration action and free-space distance annotations.

\begin{table*}[!t]\color{black}
	\small
	\centering
	\caption{Comparison of VLN dataset properties. $N_{\mathrm{traj}}$ is the number of trajectories; $\overline{\mathrm{len(traj)}}$ and $\overline{\mathrm{len(instr)}}$ are the average trajectory and instruction lengths; $\overline{N}_{\mathrm{act}}$ is the average number of actions per trajectory; and $N_{\mathrm{vocab}}$ is the vocabulary size. KinoUAV statistics refer to the language-annotated full-navigation data, excluding Auxiliary Subset.}
	\resizebox{1.0\textwidth}{!}{
		\begin{tabular}{l||c|ccccccc}
			\noalign{\hrule height 1.0pt}
			\rule{0pt}{2.8ex}\textbf{Dataset} &
			\shortstack{Kinodynamic modeling} &
			$N_{\mathrm{traj}}$ &
			$\overline{\mathrm{len(traj)}}$ &
			$\overline{\mathrm{len(instr)}}$ &
			\shortstack{Action\\space} &
			$\overline{N}_{\mathrm{act}}$ &
			Environment &
			$N_{\mathrm{vocab}}$\\
			\hline\hline
			R2R & No & 7189 & 10 & 29 & Graph & 5 & Matterport3D & 3.1K\\
			RxR & No & 13992 & 14.9 & 129 & Graph & 8 & Matterport3D & 7.0K\\
			TouchDown & No & 9326 & 313.9 & 90 & Graph & 35 & Google Street View & 5.0K\\
			VLN-CE & No & 4475 & 11.1 & 19 & 2 DoF & 56 & Matterport3D & 4.3K\\
			\hline\hline
			LANI & No & 6000 & 16 & -- & 2 DoF & 116 & CHALET & 2.3K\\
			AVDN & No & 6269 & 144.7 & 89 & 3 DoF & 7 & xView & 3.3K\\
			AerialVLN \cite{liu2023aerialvln} & No & 8446 & 661.8 & 83 & 4 DoF & 204 & AirSim & 4.5K\\
			CityNav & No & 32637 & 545 & 26 & 4 DoF & -- & SensatUrban & 6.6K\\
			TRAVEL \cite{wang2025towards} & No & 12149 & 255 & 104 & 6 DoF & 264 & AirSim & 10.8K\\
			OpenFly \cite{gao2025openfly} & No & 100K & 99.1 & 59 & 4 DoF & 35 & AirSim, GTA5, 3D GS, GE & 15.6K\\
			\hline
			\rowcolor{gray!15}KinoUAV (Ours) & Yes & 150K & 146.4 & 78 & 6 DoF & 84 & AirSim & 16.4K\\
			\noalign{\hrule height 1.0pt}
	\end{tabular}}
	\label{tab:dataset_comparison}
\end{table*}

\subsubsection{Auxiliary Subset}
Auxiliary Subset focuses on low-level waypoint-following behavior. We randomly generate waypoint chains and use hierarchical kinodynamic A$^*$ to produce the corresponding trajectories. Each valid control step is paired with the active waypoint used during trajectory generation, rather than the final endpoint of the entire chain. As the UAV progresses through the chain, the waypoint label changes accordingly, providing supervision for successive local navigation objectives. This subset contains only low-level samples $D_l^k$ and does not require language-instruction annotations. It complements the complete navigation trajectories in Main Subset by introducing varied waypoint transitions for training the low-level controller.

\subsubsection{Main Subset}
Main Subset contains complete target-oriented navigation trajectories generated using hierarchical kinodynamic A$^*$. It supports both high-level waypoint planning and low-level acceleration control. The high-level samples cover Cruise, where the UAV approaches the target region using spatial guidance, and Seek, where it visually locates the target using its description. Together with the recorded UAV states, images, acceleration actions, and DFDs, these annotations form the low-level samples $D_l^k$.

For Cruise-stage samples, we traverse each trajectory in temporal order. From the current state at step $k_i$, we identify the first subsequent state whose distance from $\mathbf{p}^{k_i}$ exceeds $10\,\mathrm{m}$. Its displacement is transformed into the current UAV body frame and scaled to $10\,\mathrm{m}$ while preserving its direction. If the trajectory endpoint is less than $10\,\mathrm{m}$ away, its body-frame displacement is used without scaling. The instruction $T^i$ provides the target's relative direction and distance from the current UAV position.

The target description is generated based on four images: two top-down views captured from $5\,\mathrm{m}$ and $8\,\mathrm{m}$ directly above the target, and two frontal views captured from a height of $2\,\mathrm{m}$ while moving along horizontal circular paths centered on the target. For the frontal views, the camera is positioned at a height of $2\,\mathrm{m}$ on circular paths with radii of $5\,\mathrm{m}$ meters and $8\,\mathrm{m}$ meters, respectively, with the camera always facing the target. These images are fed into Qwen3-VL-235B to generate the target description. Additionally, each trajectory is assigned one of nine application scenarios—such as medical rescue or logistics transport—to enhance instruction diversity. We require the large model to incorporate the specified application context when describing the characteristics of a target.

For samples in the "Seek" phase, ten points located more than $5\,\mathrm{m}$ meters above the target are randomly sampled. Points are then filtered based on the condition that the top-down view's DFD value is not less than the point's actual distance to the target, ensuring the target is unobstructed in the image. Waypoint labels represent the target's coordinates in the vehicle's local coordinate system; if the magnitude of these coordinates exceeds 10 meters, the vector is scaled down to a length of $10\,\mathrm{m}$ meters. Instructions for the "Seek" phase consist solely of the generated target description, necessitating visual localization based on the input images.

To construct the dataset, we first reserve one of the 16 scenes exclusively for Test Unseen Map. In the remaining 15 scenes, we partition the target objects into disjoint seen and unseen sets. Target-object placement is required only for Main Subset. For each trajectory in this subset, we first select the start and endpoint positions and place a target object at the endpoint. Seen objects are used for the training set and Test Seen, whereas unseen objects are used for Test Unseen Object. We then plan the trajectory using hierarchical kinodynamic A$^*$ and collect the corresponding observations and annotations following the procedures described above. For Auxiliary Subset, we select a start position without placing a target object, randomly generate a waypoint chain, and collect the resulting trajectory and low-level annotations. Auxiliary Subset is constructed entirely within the remaining 15 scenes and is used only for training. Finally, we apply the Main Subset collection procedure to the held-out scene to construct Test Unseen Map. The resulting dataset comprises 147K training trajectories and 1K trajectories for each of Test Seen, Test Unseen Object, and Test Unseen Map, while all 150K trajectories in Auxiliary Subset are used for training.

\begin{figure}[t]
	\centering
	\includegraphics[width=\columnwidth]{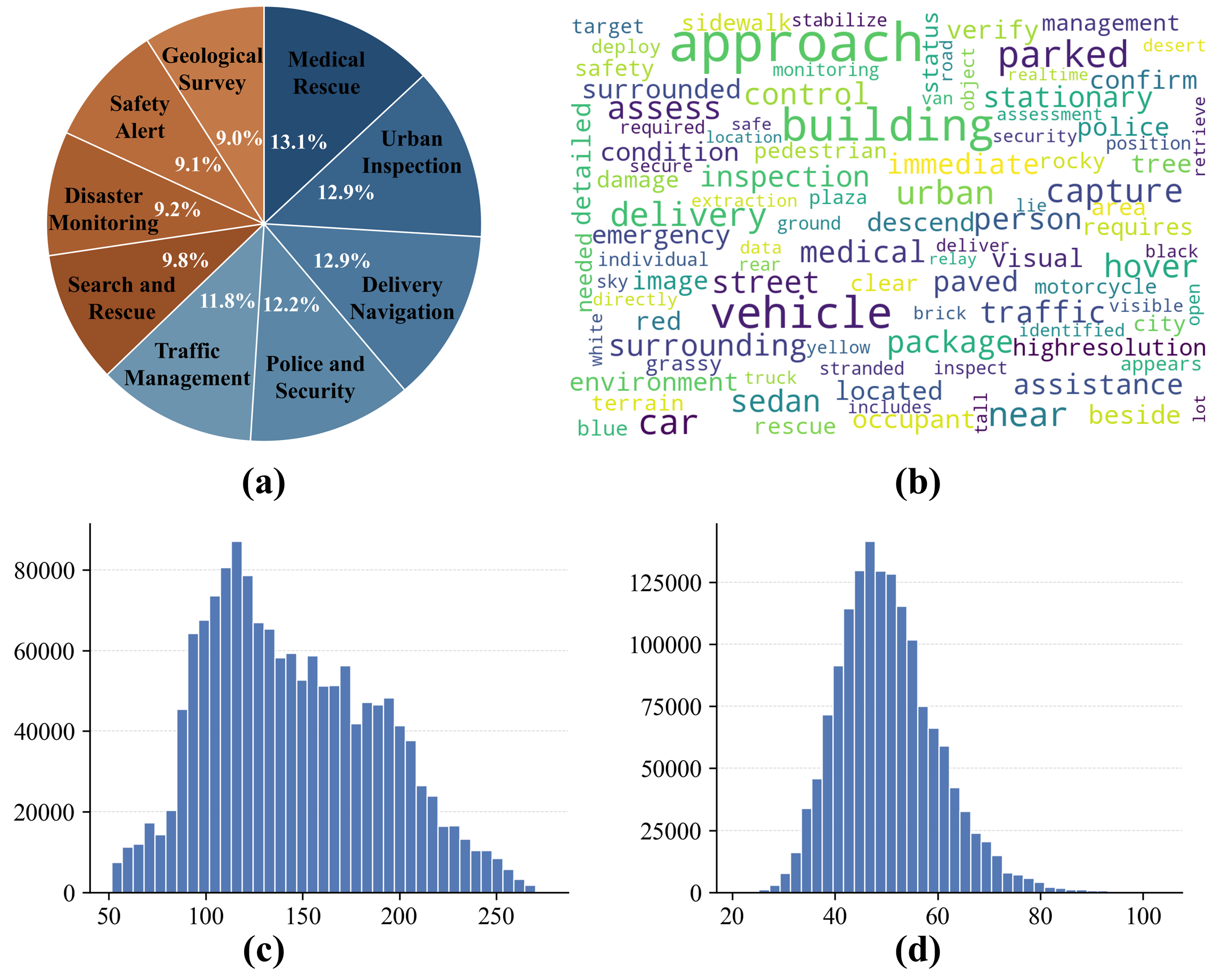}
	\caption{Dataset statistics. (a) Scenario categories used to annotate target descriptions; blue denotes urban scenarios and orange denotes non-urban scenarios. (b) Word cloud of the annotated descriptions. (c) and (d) Trajectory length distributions of Main Subset and Auxiliary Subset, respectively.}
	\label{fig:data_analysis}
\end{figure}

Fig.~\ref{fig:data_analysis} summarizes the resulting dataset. Panel (a) shows the nine annotation scenarios, with urban scenarios accounting for $62.9\%$ of the samples, while panel (b) illustrates the vocabulary of the generated target descriptions. Panels (c) and (d) show the distribution of trajectory lengths.

\begin{figure}[!h]
	\centering
	\includegraphics[width=\columnwidth]{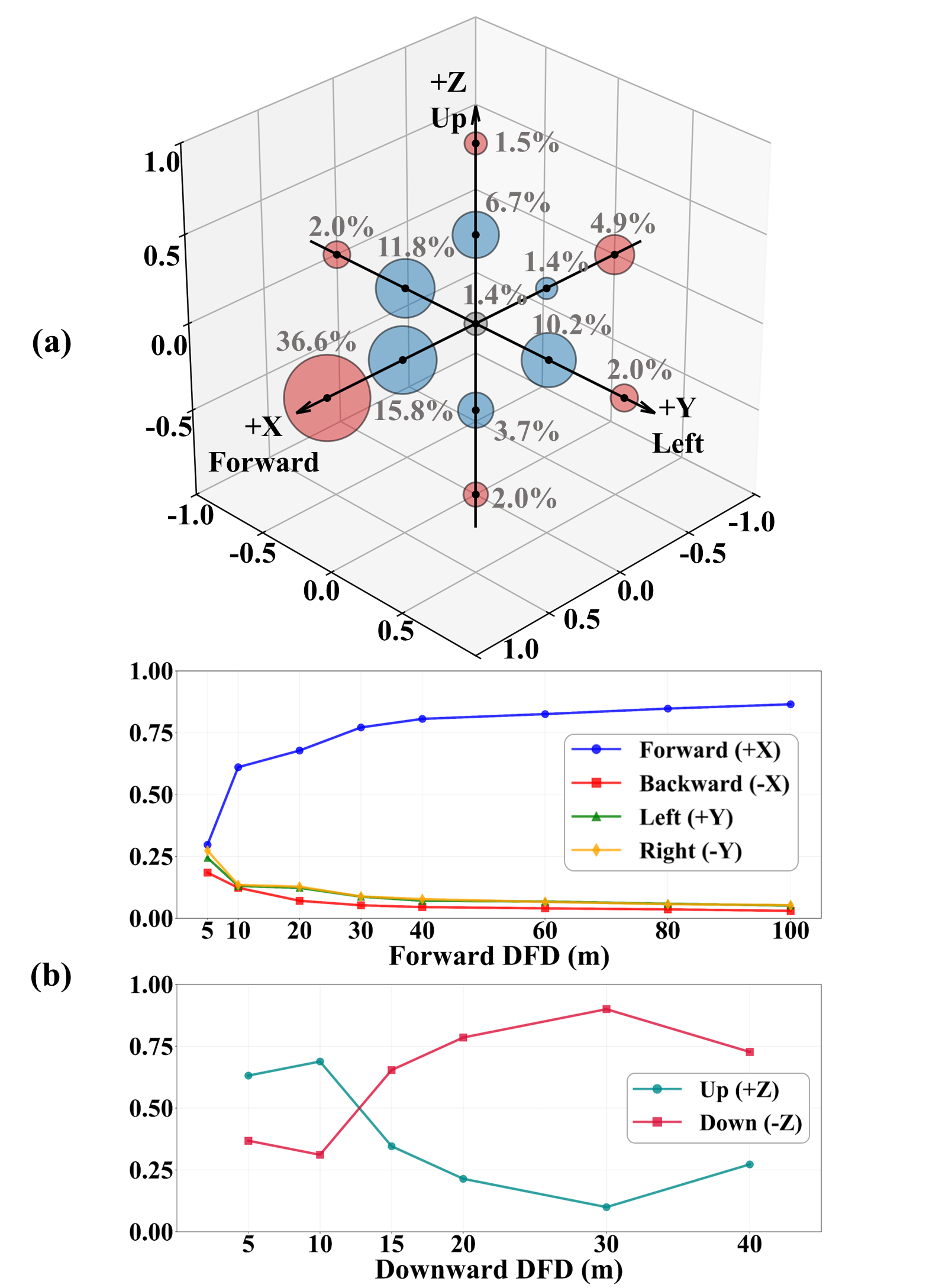}
	\caption{Acceleration action statistics in our dataset. (a) Overall action distribution. (b) Action probabilities versus DFD. Actions are grouped by direction regardless of magnitude, with the infrequent stay action excluded.}
	\label{fig:dfd_analysis}
\end{figure}

\begin{figure*}[t]
	\centering
	\includegraphics[width=\textwidth]{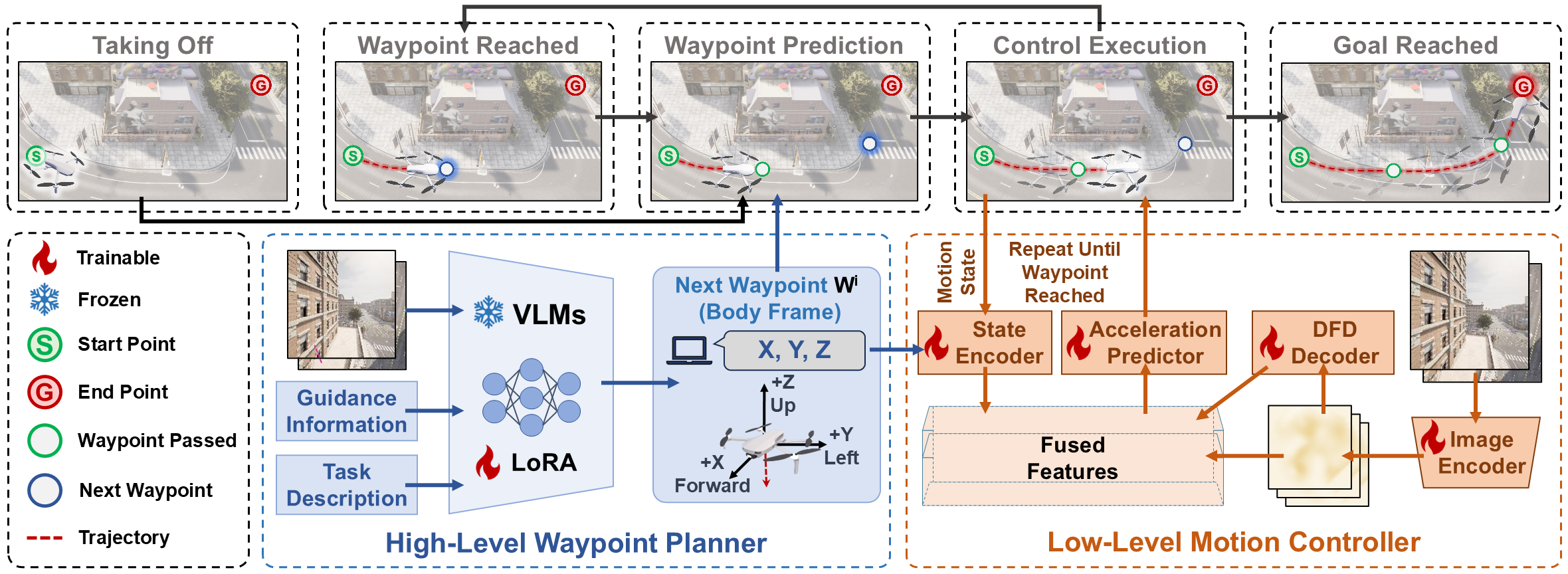}
	\caption{Overview of KinoUAV-Agent. It hierarchically integrates waypoint planning, DFD estimation, and acceleration-based motion control.}
	
	\label{fig:Model}
\end{figure*}

Table~\ref{tab:dataset_comparison} compares how different VLN datasets are constructed and their main features. KinoUAV uses automatic trajectory collection and instruction annotation. OpenFly also automates these steps, but KinoUAV further includes kinodynamic modeling in trajectory generation to account for UAV motion constraints. This allows large-scale collection of trajectory data for full navigation and waypoint tracking.

Fig.~\ref{fig:dfd_analysis}(a) shows the distribution of acceleration actions in KinoUAV, with larger spheres indicating more frequent actions. Fig.~\ref{fig:dfd_analysis}(b) shows how action selection varies with DFD. The forward action becomes less likely as the forward DFD decreases. The same trend is observed for the downward action when the bottom DFD decreases. These results suggest that the UAV is less likely to move toward directions with limited free space.

\subsection{KinoUAV-Agent}

As shown in Fig.~\ref{fig:Model}, KinoUAV-Agent adopts a hierarchical
navigation framework comprising a high-level waypoint planner and a
low-level motion controller. The two modules operate at different
frequencies: the high-level planner generates a new waypoint when a
planning event occurs, whereas the low-level controller selects an
acceleration action at every control step. We use $i$ to index
waypoints and $k$ to index control steps.

The initial UAV position is
$\mathbf{p}^{0}={p}_{S}$. Before invoking the high-level planner,
we assign the takeoff waypoint
$\mathbf{w}^{0}=(0,0,z_H)^\top$, where $z_H$ is the prescribed cruise
altitude. The low-level controller first guides the UAV from the
ground to this waypoint. Once the takeoff waypoint is reached, the
high-level planner begins generating waypoints for target-oriented
navigation.

\begin{figure*}[t]
	\centering
	\includegraphics[width=\textwidth]{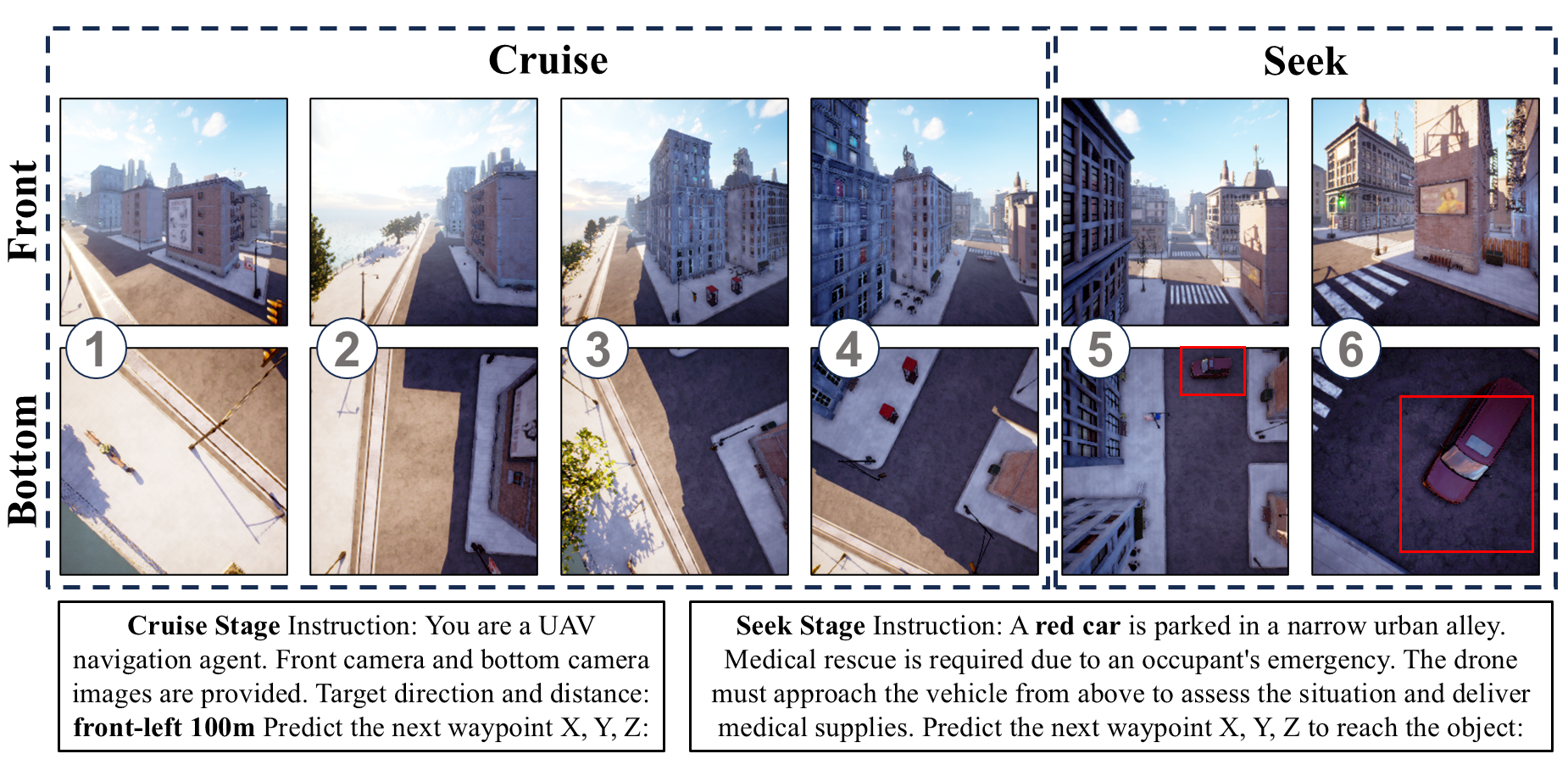}
	\caption{Illustration of the Cruise--Seek navigation process and stage-specific instructions. Front- and bottom-view observations are shown at six successive navigation moments. During Cruise (1--4), the UAV approaches the target region using direction and distance guidance. During Seek (5--6), it visually localizes and approaches the target using its textual description. Red boxes highlight the target, and representative instructions for both stages are shown below.}
	\label{fig:full_navigation}
\end{figure*}

As shown in Fig.~\ref{fig:full_navigation}, the subsequent flight consists of Cruise and Seek stages. During
Cruise, coarse directional guidance helps the UAV move toward the
vicinity of the target. During Seek, the UAV uses the target
description and its visual observations to locate and approach the
object. Within either stage, the low-level controller follows the
current waypoint while responding to obstacles observed in the front
and bottom views. When the distance to the current waypoint is less than $r_{\mathrm{arr}}$, the high-level planner
uses the latest observations to generate the next one. A stage
transition can also trigger immediate replanning, so the UAV need
not finish a previously assigned Cruise waypoint after entering
the Seek region. This cycle continues until the navigation task
terminates or the maximum number of control steps is reached.

\begin{algorithm}[t]
	\caption{Closed-Loop Navigation of KinoUAV-Agent}
	\label{alg:kinouav_navigation}
	\footnotesize
	\KwIn{Start $\mathbf{p}_{\mathrm{s}}$, goal $\mathbf{p}_{\mathrm{g}}$, target description $T_{\mathrm{obj}}$, takeoff height $z_H$, thresholds $r_{\mathrm{arr}}$ and $r_{\mathrm{stage}}$, action set $\mathcal{A}_{\mathrm{UAV}}$, maximum steps $K_{\max}$}
	\KwOut{Executed trajectory $\mathcal{T}$}
	
	$\mathbf{p}^0\gets{p}_{S}$; $\mathbf{W}^0\gets(0,0,z_H)^\top$; $\mathbf{g}_H\gets\mathbf{p}_{\mathrm{g}}+(0,0,z_H)^\top$\;
	$i\gets0$; $k\gets0$; stage $\gets\mathrm{Cruise}$\;
	Initialize $\mathcal{T}$ with the initial UAV state\;
	\While{$k<K_{\max}$ and the task is not terminated}{
		Observe $I_f^k$, $I_b^k$, and the current UAV state\;
		$\mathrm{update}\gets\mathrm{false}$\;
		
		\eIf{stage $=\mathrm{Cruise}$ and $\|\mathbf{p}^k-\mathbf{g}_H\|\leq r_{\mathrm{stage}}$}{
			stage $\gets\mathrm{Seek}$\;
			$\mathrm{update}\gets\mathrm{true}$\;
		}{
			\If{$\|\mathbf{p}^k-\mathbf{W}^i\|\leq r_{\mathrm{arr}}$}{
				$\mathrm{update}\gets\mathrm{true}$\;
			}
		}
		
		\If{$\mathrm{update}=\mathrm{true}$}{
			$i\gets i+1$\;
			\eIf{stage $=\mathrm{Cruise}$}{
				$T^i\gets$ current direction and distance guidance\;
			}{
				$T^i\gets T_{\mathrm{obj}}$\;
			}
			
			$\mathbf{w}^{i}\gets f_h\!\left(I_f^k,I_b^k,T^i\right)$\;
			
			$\mathbf{W}^i\gets\mathbf{p}^k+\mathbf{R}^{k}\mathbf{w}^{i}$\;
		}
		
		Compute $\mathbf{s}_i^k$ from the current UAV state and $\mathbf{W}^i$\;
		
		$\mathbf{a}_{\mathrm{UAV, raw}}^k\gets f_l\!\left(I_f^k,I_b^k,\mathbf{s}_i^k\right)$\;
		Select the nearest acceleration in $\mathcal{A}_{\mathrm{UAV}}$ to $\mathbf{a}_{\mathrm{UAV,raw}}^k$\;
		Generate the next reference point and track it\;
		Append the updated UAV state to $\mathcal{T}$\;
		$k\gets k+1$\;
	}
	\Return $\mathcal{T}$\;
\end{algorithm}

\subsubsection{High-Level Waypoint Planner}

Let $\mathbf{g}=(x_g,y_g,0)^\top$ denote the ground-level target
position in the coordinate frame defined above. We define the
reference point for the Cruise stage as
$\mathbf{g}_H=(x_g,y_g,z_H)^\top$, vertically above the target at
cruise altitude. This point defines the vicinity in which the
navigation stage changes; its exact coordinate is not provided as
an input to the high-level planner.

At a planning step $k$, the high-level planner receives the
front-view image $I_f^k$, bottom-view image $I_b^k$, and a
stage-dependent textual cue $T^i$. It predicts the next waypoint as
\begin{equation}
	\mathbf{w}^{i}
	=f_h\!\left(I_f^k,I_b^k,T^i\right),
	\qquad i\geq 1,
	\label{eq:high_level_planner}
\end{equation}
The high-level module consistently outputs the next waypoint $\mathbf{w}^{i}$ in the body frame, with the corresponding waypoint in the world frame being $\mathbf{W}^{i} = \mathbf{p}^{k}+\mathbf{R}^{k}\mathbf{w}^{i}$, where $\mathbf{R}^{k}$ is the rotation matrix from the body frame to the world frame at the waypoint-generation step $k$. The same planner $f_h$ is used in both navigation stages, while
the content of $T^i$ changes with the stage.

During Cruise, $T^i=T_{\mathrm{dir}}^i$ specifies a discrete
direction and distance relative to the UAV, such as \textit{front-left} 50m, \textit{right} 32m. The planner combines this coarse cue with
the current visual observations to select successive waypoints.
The UAV remains in Cruise while it is outside the target region.
When
\begin{equation}
	\|\mathbf{p}^{k}-\mathbf{g}_H\|
	\leq r_{\mathrm{stage}},
	\label{eq:seek_transition}
\end{equation}
the navigation process switches to Seek, where
$r_{\mathrm{stage}}$ is the target-region radius.

In Seek, the directional cue is removed and replaced by the
appearance description $T_{\mathrm{obj}}$ of the ground-level
target. The planner therefore receives
$(I_f^k,I_b^k,T_{\mathrm{obj}})$ and predicts waypoints that
progressively bring the UAV closer to the visually identified
object. If the stage changes while the UAV is following a Cruise
waypoint, that waypoint is discarded and a new Seek waypoint is
generated from the current observations. Subsequent Seek waypoints
are generated whenever the current one is reached.

\subsubsection{Low-Level Motion Controller}

For an active waypoint $\mathbf{W}^{i}$, the low-level controller
operates at each control step $k$. It takes the two camera views
and a waypoint-relative state vector as input, and predicts the
control acceleration
\begin{equation}
	\mathbf{a}_{\mathrm{UAV, raw}}^k
	=f_l\!\left(I_f^k,I_b^k,\mathbf{s}_i^k\right).
	\label{eq:low_level_controller}
\end{equation}
This is the active control acceleration in the kinodynamic model;
the total acceleration additionally includes velocity-dependent
drag.

The waypoint-relative distance and direction are
\begin{equation}
	\begin{aligned}
		r_i^k
		&=\|\mathbf{W}^{i}-\mathbf{p}^{k}\|,\\
		\mathbf{D}_i^k
		&=\frac{\mathbf{W}^{i}-\mathbf{p}^{k}}{r_i^k},
		\qquad r_i^k>0.
	\end{aligned}
	\label{eq:waypoint_relative_state}
\end{equation}
We set $\mathbf{D}_i^k=\mathbf{0}$ when $r_i^k=0$. Let
$v_h^k=\sqrt{(v_x^k)^2+(v_y^k)^2}$ and $v_v^k=v_z^k$
denote the horizontal speed and vertical velocity, respectively.
The state supplied to the controller is
\begin{equation}
	\mathbf{s}_i^k
	=
	\left[
	\mathbf{D}_i^k,\,
	g_{\mathrm{dist}}(r_i^k),\,
	v_h^k,\,
	v_v^k
	\right]^\top,
	\label{eq:controller_state}
\end{equation}
where $g_{\mathrm{dist}}$ applies logarithmic distance normalization:
\begin{equation}
	g_{\mathrm{dist}}(r)
	=
	\frac{1}{2}\min\left\{
	\log_{10}\!\left(\max\{r,10^{-2}\}\right),\,2
	\right\}.
	\label{eq:distance_normalization}
\end{equation}
The state encodes the waypoint direction, normalized distance,
and the UAV's horizontal and vertical velocities.

The low-level controller comprises a shared visual encoder
$f_{\mathrm{vision}}$, a DFD decoder $f_{\mathrm{DFD}}$, and an
acceleration predictor $f_{\mathrm{action}}$. The encoder extracts
features from both views:
\begin{equation}
	\mathbf{f}_{\mathrm{img}}^k
	=f_{\mathrm{vision}}\!\left(I_f^k,I_b^k\right).
	\label{eq:visual_features}
\end{equation}
The visual encoder uses a ResNet18 backbone with the final classification layer removed and the global average pooling layer retained. The front- and bottom-view images are processed separately using shared backbone weights, producing a 512-dimensional feature vector for each view. These vectors are concatenated to form the joint visual representation $\mathbf{f}_{\mathrm{img}}^k \in \mathbb{R}^{1024}$. The DFD decoder estimates the front- and bottom-view
directional free-space distance representations:
\begin{equation}
	\mathbf{d}^k
	=f_{\mathrm{DFD}}\!\left(
	\mathbf{f}_{\mathrm{img}}^k\right),
	\qquad
	\mathbf{d}^k
	=[\mathbf{d}_f^k,\mathbf{d}_b^k].
	\label{eq:dfd_prediction}
\end{equation}
The acceleration predictor then combines the visual features, waypoint-relative state, and estimated DFDs:
\begin{equation}
	\mathbf{a}_{\mathrm{UAV, raw}}^k
	=f_{\mathrm{action}}\!\left(
	\mathbf{f}_{\mathrm{img}}^k,
	\mathbf{s}_i^k,
	\mathbf{d}^k
	\right).
	\label{eq:acceleration_prediction}
\end{equation}

The acceleration predicted in (\ref{eq:acceleration_prediction}) is expressed in the UAV body frame. Before execution, we project it onto the discrete action set $\mathcal{A}_{\mathrm{UAV}}$. The executed body-frame action is the prototype nearest to the network prediction in Euclidean distance:
\begin{equation}
	\mathbf{a}_{\mathrm{UAV}}^k
	=\underset{\mathbf{u}\in\mathcal{A}_{\mathrm{UAV}}}{\operatorname{arg\,min}}
	\left\|\mathbf{u}-\mathbf{a}_{\mathrm{UAV, raw}}^k\right\|_2.
\end{equation}
We rotate this action into the inertial frame and add the velocity-dependent drag acceleration:
\begin{equation}
	\mathbf{a}_{\mathrm{ref}}^k
	=\mathbf{R}^{k}\mathbf{a}_{\mathrm{UAV}}^k
	-k_1\mathbf{v}^k
	-k_2\|\mathbf{v}^k\|\mathbf{v}^k,
\end{equation}
Starting from the current realized position and velocity, we compute the next reference position $\mathbf{p}_{\mathrm{ref}}^{k+1}$ as
\begin{equation}
	\begin{aligned}
		\mathbf{v}_{\mathrm{ref}}^{k+1}
		&=\mathbf{v}^k+\mathbf{a}_{\mathrm{ref}}^k\Delta t,\\
		\mathbf{p}_{\mathrm{ref}}^{k+1}
		&=\mathbf{p}^k+\mathbf{v}_{\mathrm{ref}}^{k+1}\Delta t.
	\end{aligned}
\end{equation}
The next reference position is tracked through the flight-control API. The realized UAV state is then read from the simulator to update the waypoint-relative state. After the action is executed, the UAV state and its distance to the active waypoint are updated. The same waypoint remains active while $r_i^{k+1}>r_{\mathrm{arr}}$. Once $r_i^{k+1}\leq r_{\mathrm{arr}}$, the waypoint is considered reached and the high-level planner is invoked again, unless the navigation task has already terminated. Here $r_{\mathrm{arr}}$ is the waypoint arrival radius, distinct from the larger target-region radius $r_{\mathrm{stage}}$ used to switch from Cruise to Seek. The closed-loop interaction between the two modules is illustrated in Fig.~\ref{fig:Model}.

For training, we supervise both the control acceleration and the DFD prediction using mean squared error (MSE). Let $\mathbf{a}_{\mathrm{UAV,gt}}^k$ and $\mathbf{d}_{\mathrm{gt}}^k$ denote their ground-truth labels. The losses are
\begin{equation}
	\begin{aligned}
		\mathcal{L}_{\mathrm{action}}
		&=\operatorname{MSE}\!\left(\mathbf{a}_{\mathrm{UAV, raw}}^k,\mathbf{a}_{\mathrm{UAV,gt}}^k\right),\\
		\mathcal{L}_{\mathrm{dist}}
		&=\operatorname{MSE}\!\left(\mathbf{d}^k,g_{\mathrm{dist}}(\mathbf{d}_{\mathrm{gt}}^k)\right),\\
		\mathcal{L}
		&=\lambda_{\mathrm{action}}\mathcal{L}_{\mathrm{action}}
		+\lambda_{\mathrm{dist}}\mathcal{L}_{\mathrm{dist}},
	\end{aligned}
\end{equation}
where MSE averages the squared errors over all elements in a training batch, and $\lambda_{\mathrm{action}}$ and $\lambda_{\mathrm{dist}}$ weight the two objectives.

\section{Experiments}

\subsection{Implementation Details}

All experiments were conducted on Ubuntu 22.04 with Python 3.10 and PyTorch 2.8.0. Both model training and inference used eight NVIDIA RTX 5090 GPUs.

\subsubsection{Hyperparameter Settings}

For dataset construction, we set the UAV mass to $m_{\mathrm{UAV}}=1\,\mathrm{kg}$ and the maximum gravity-compensated control force to $F_{\mathrm{UAV,max}}=1\,\mathrm{N}$. The mass-normalized linear and quadratic drag coefficients are $k_1=0.2\,\mathrm{s}^{-1}$ and $k_2=0.01\,\mathrm{m}^{-1}$, respectively. The cost weights are $(\lambda_1,\lambda_2,\lambda_3,\lambda_4)=(1,2,2,0.5)$, and the waypoint arrival radius is $r_{\mathrm{arr}}=3\,\mathrm{m}$. Simulation time interval $\Delta t=1s$.

We fine-tune all the vision-language models with LoRA rank 8, targeting all eligible modules, using a learning rate of $1.0\times10^{-4}$ for 2 epochs. For the low-level controller, the action and DFD loss weights are $\lambda_{\mathrm{action}}=0.67$ and $\lambda_{\mathrm{dist}}=0.33$, respectively. During evaluation, the cruise altitude is $z_H=12\,\mathrm{m}$ and the stage-transition threshold is $r_{\mathrm{stage}}=5\,\mathrm{m}$. The image resolution used for both model training and inference is 256×256. The low-level controller is trained for 50 epochs with a learning rate of $1.0\times10^{-4}$ and a batch size of 128. For low-level supervision on Main Subset, we use the trajectory
endpoint as the waypoint at each control step, whereas Auxiliary Subset
uses the active waypoint from the sampled chain.

\begin{table*}[!h]\color{black}
	\small
	\centering
	\caption{Navigation performance on KinoUAV Test Seen. SR, OSR, NE, SPL, and CR denote success rate, oracle success rate, navigation error, success weighted by path length, and collision rate, respectively. Mean speed and mean absolute yaw rate are reported in $\mathrm{m/s}$ and $\mathrm{rad/s}$, respectively.}
	\begin{tabular}{l||ccccc|cc}
		\noalign{\hrule height 1.0pt}
		\rule{0pt}{2.8ex}\textbf{Method} &
		\textbf{SR} (\%)$\uparrow$ &
		\textbf{OSR} (\%)$\uparrow$ &
		\textbf{NE} (m)$\downarrow$ &
		\textbf{SPL} (\%)$\uparrow$ &
		\textbf{CR} (\%)$\downarrow$ &
		\textbf{$\overline{v}$}(m/s)$\uparrow$ &
		\textbf{$\overline{|\omega_z^k|}$}(rad/s)$\downarrow$\\
		\hline\hline
		\rowcolor{gray!15}\multicolumn{8}{c}{\textbf{Existing Method}}\\
		\hline
		Random & 0.0 & 0.2 & 138.0 & 0.0 & 42.5 & 0.588 & 0.215\\
		CMA \cite{anderson2018vision} & 11.5 & 13.6 & 86.4 & 9.6 & 34.5 & 0.811 & 0.096\\
		TravelUAV \cite{wang2025towards} & 20.1 & 53.9 & 55.9 & 16.4 & 16.9 & 0.917 & 0.100\\
		AerialVLA \cite{xu2026aerialvla} & 52.3 & 55.0 & 32.0 & 36.9 & 28.3 & 0.648 & 0.113\\
		\hline\hline
		\rowcolor{gray!15}\multicolumn{8}{c}{\textbf{Zero-shot VLM Backbones on KinoUAV}}\\
		\hline
		Qwen3VL-2B & 0.2 & 1.0 & 125.4 & 0.1 & 23.0 & 1.275 & 0.040\\
		Qwen3VL-4B & 1.8 & 42.3 & 89.0 & 0.3 & 43.3 & 1.318 & 0.078\\
		Qwen3VL-8B & 3.7 & 53.6 & 89.4 & 2.9 & 32.8 & 1.248 & 0.081\\
		\hline\hline
		\rowcolor{gray!15}\multicolumn{8}{c}{\textbf{Fine-tuned VLM Backbones on KinoUAV}}\\
		\hline
		InternVL3-2B & 75.8 & 78.9 & 19.0 & 60.2 & 14.5 & 1.602 & 0.042\\
		InternVL3-8B & 78.5 & 81.1 & 9.8 & 62.4 & 12.8 & 1.606 & 0.051\\
		LFM2.5VL-1.6B & 78.3 & 80.7 & 12.6 & 62.1 & 14.4 & 1.686 & 0.046\\
		LLaVA1.5-7B & 83.6 & 84.6 & 15.0 & 67.3 & 9.2 & \textbf{1.790} & 0.041\\
		Qwen3VL-2B & 83.6 & 86.8 & 16.2 & 67.4 & 7.0 & 1.582 & \textbf{0.035}\\
		Qwen3VL-4B & 87.3 & 89.8 & 10.0 & 71.3 & 6.5 & 1.675 & 0.044\\
		\rowcolor{gray!15}\textbf{Qwen3VL-8B (KinoUAV-Agent)} & \textbf{89.3} & \textbf{90.8} & \textbf{8.8} & \textbf{71.9} & \textbf{6.0} & 1.680 & 0.045\\
		\noalign{\hrule height 1.0pt}
	\end{tabular}
	\label{tab:Test_Seen_comparison}
\end{table*}

\begin{table*}[!t]\color{black}
	\small
	\centering
	\caption{Navigation performance and selected kinematic statistics on KinoUAV Test Unseen Object and Test Unseen Map.}
	\begin{tabular}{l||ccccc|cc}
		\noalign{\hrule height 1.0pt}
		\rule{0pt}{2.8ex}\textbf{Method} &
		\textbf{SR} (\%)$\uparrow$ &
		\textbf{OSR} (\%)$\uparrow$ &
		\textbf{NE} (m)$\downarrow$ &
		\textbf{SPL} (\%)$\uparrow$ &
		\textbf{CR} (\%)$\downarrow$ &
		\textbf{$\overline{v}$}(m/s)$\uparrow$ &
		\textbf{$\overline{|\omega_z^k|}$}(rad/s)$\downarrow$\\
		\hline\hline
		\rowcolor{gray!15}\multicolumn{8}{c}{\textbf{Test Unseen Object}}\\
		\hline
		Random & 0.0 & 0.0 & 152.1 & 0.0 & 32.8 & 0.634 & 0.189\\
		CMA \cite{anderson2018vision} & 9.6 & 13.6 & 84.1 & 7.9 & 30.1 & 0.943 & 0.115\\
		TravelUAV \cite{wang2025towards} & 21.4 & 52.5 & 56.8 & 16.9 & 17.2 & 0.902 & 0.098\\
		AerialVLA \cite{xu2026aerialvla} & 51.5 & 56.3 & 31.2 & 37.6 & 27.9 & 0.661 & 0.116\\
		\rowcolor{gray!15}\textbf{KinoUAV-Agent} & \textbf{87.1} & \textbf{89.7} & \textbf{8.6} & \textbf{72.4} & \textbf{6.2} & 1.652 & 0.044\\
		\hline\hline
		\rowcolor{gray!15}\multicolumn{8}{c}{\textbf{Test Unseen Map}}\\
		\hline
		Random & 0.1 & 0.3 & 139.5 & 0 & 41.8 & 0.601 & 0.209\\
		CMA \cite{anderson2018vision} & 3.4 & 6.1 & 105.8 & 1.9 & 32.4 & 0.928 & 0.121\\
		TravelUAV \cite{wang2025towards} & 12.7 & 31.2 & 68.2 & 9.1 & 24.5 & 0.931 & 0.103\\
		AerialVLA \cite{xu2026aerialvla} & 39.4 & 47 & 40.7 & 26.2 & 33.7 & 0.637 & 0.11\\
		\rowcolor{gray!15}\textbf{KinoUAV-Agent} & \textbf{64.6} & \textbf{75.4} & \textbf{17.1} & \textbf{51.3} & \textbf{8.9} & 1.703 & 0.046\\
		\noalign{\hrule height 1.0pt}
	\end{tabular}
	\label{tab:Test_Unseen_comparison}
\end{table*}

\subsection{Evaluation Metric}

We evaluate $N$ navigation episodes. For episode $n$, let $\mathbf{g}_n$ be the target position, and let $\mathbf{p}_n^k$ and $\mathbf{v}_n^k$ be the UAV position and velocity at step $k$, with $K_n$ denoting the final step. Success Rate (SR) requires the UAV to finish within $6\,\mathrm{m}$ of the target:

\begin{equation}
	\mathrm{SR}=\frac{1}{N}\sum_{n=1}^{N}\mathbb{I}\!\left(\|\mathbf{p}_n^{K_n}-\mathbf{g}_n\|<6\,\mathrm{m}\right),
\end{equation}
where $\mathbb{I}(\cdot)$ is the indicator function. The Oracle Success Rate (OSR) is calculated as follows: a sample is considered an "Oracle success" if any point along the drone's flight trajectory is within $20\,\mathrm{m}$ of the target, and the OSR represents the proportion of such samples. Navigation Error (NE) is the mean distance from the final UAV position to the target:
\begin{equation}
	\begin{aligned}
		\mathrm{OSR}&=\frac{1}{N}\sum_{n=1}^{N}\mathbb{I}\!\left(\min_{0\leq k\leq K_n}\|\mathbf{p}_n^k-\mathbf{g}_n\|<20\,\mathrm{m}\right),\\
		\mathrm{NE}&=\frac{1}{N}\sum_{n=1}^{N}\|\mathbf{p}_n^{K_n}-\mathbf{g}_n\|.
	\end{aligned}
\end{equation}

To jointly evaluate navigation success and path efficiency, we compute success weighted by path length (SPL). For episode $n$, let $S_n$ denote the binary success indicator defined by the SR criterion, and let $\ell_n$ and $L_n$ denote the ground-truth and executed trajectory lengths, respectively. SPL is averaged over all $N$ episodes:

\begin{equation}
	\mathrm{SPL}
	=\frac{1}{N}\sum_{n=1}^{N}
	S_n\frac{\ell_n}{\max(\ell_n,L_n)}
\end{equation}

An unsuccessful episode contributes zero to SPL. The collision rate (CR) refers to the proportion of trajectories that result in a collision. In addition, we evaluate motion characteristics using average speed and the average absolute yaw rate $\overline{|\omega_z^k|}$. These statistics reflect flight speed and changes in heading.
The yaw rate is the change in yaw divided by $\Delta t$:
\begin{equation}
	\omega_z^{k+1}
	=\frac{\mathrm{yaw}^{k+1} - \mathrm{yaw}^{k}}{\Delta t},
\end{equation}

\subsection{Comparison Experiments}

We compare KinoUAV-Agent with Random Action and three representative learned navigation baselines: CMA, TravelUAV, and AerialVLA. For a fair comparison, all three learned baselines are retrained on the KinoUAV training set and evaluated in the same simulation environments using the same test splits and evaluation criteria. Random Action randomly selects a valid acceleration action at each step until the UAV reaches the target or the maximum number of evaluation steps is reached.

All learned baselines are retrained on our dataset. For CMA \cite{anderson2018vision}, we adapt its LSTM to use concatenated front- and bottom-view image features and replace discrete action classification with three-dimensional acceleration regression. We train the model for 2 epochs with a batch size of 16 and a learning rate of $10^{-4}$. For TravelUAV \cite{wang2025towards}, we convert the input format while retaining the original model architecture and training configuration. For AerialVLA \cite{xu2026aerialvla}, we convert the next trajectory point in each training sample into the three quantized action components required by the model. We retain its original training settings, except that the number of training epochs is set to 2.

To assess the applicability of different vision-language models (VLMs) to our high-level planner, we fine-tune Qwen3-VL-2B/4B/8B-Instruct \cite{Qwen3-VL}, InternVL3-2B/8B \cite{zhu2025internvl3}, LFM2.5-VL-1.6B \cite{liquidai2025lfm2}, and LLaVA-1.5-7B \cite{liu2024improved}. All VLMs use the same dataset splits and inference pipeline. We additionally evaluate the three Qwen3-VL backbones in the zero-shot setting to examine the contribution of task-specific fine-tuning. In zero-shot testing, only the low-level modules are trained.

We evaluate navigation performance on KinoUAV Test Seen, Test Unseen Object, and Test Unseen Map. SR, OSR, NE, and SPL measure task completion, proximity to the target, and path efficiency. We additionally report CR, mean speed, and mean absolute yaw rate to characterize collision avoidance and flight behavior. These complementary metrics allow us to assess whether improved navigation success is accompanied by efficient motion and reduced heading variation.

\begin{table*}[!t]\color{black}
	\small
	\centering
	\caption{Ablation study on training subsets and data scale on KinoUAV Test Seen. The training fraction is applied to each selected subset independently. All variants use the same supervised DFD branch and navigation pipeline.}
	\begin{tabular}{lc||ccccc|cc}
		\noalign{\hrule height 1.0pt}
		\rule{0pt}{2.8ex}\textbf{Subsets} &
		\textbf{Fraction} &
		\textbf{SR} (\%)$\uparrow$ &
		\textbf{OSR} (\%)$\uparrow$ &
		\textbf{NE} (m)$\downarrow$ &
		\textbf{SPL} (\%)$\uparrow$ &
		\textbf{CR} (\%)$\downarrow$ &
		\textbf{$\overline{v}$}(m/s)$\uparrow$ &
		\textbf{$\overline{|\omega_z^k|}$}(rad/s)$\downarrow$\\
		\hline\hline
		Only Main Subset & -- & 71.2 & 77.1 & 25.9 & 55.2 & 15.4 & 1.865 & 0.045\\
		\hline
		All & 1\%   & 0.9  & 10.5 & 170.4 & 0.2 & 68.5 & 1.583 & 0.023\\
		All & 10\%  & 74.2 & 77.6 & 18.4  & 60.1 & 17.0 & 1.664 & 0.032\\
		\rowcolor{gray!15}\textbf{All} & \textbf{100\%} &
		\textbf{89.3} & \textbf{90.8} & \textbf{8.8} &
		\textbf{71.9} & \textbf{6.0} & 1.680 & 0.045\\
		\noalign{\hrule height 1.0pt}
	\end{tabular}
	\label{tab:dataset_ablation}
\end{table*}

\begin{table*}[!t]\color{black}
	\small
	\centering
	\caption{Ablation of navigation and control configurations on KinoUAV Test Seen. The full configuration uses Cruise--Seek guidance, initializes the UAV heading toward the next ground-truth waypoint, and projects predicted accelerations onto $\mathcal{A}_{\mathrm{UAV}}$. Each variant modifies the indicated setting while retaining the remaining settings of the full configuration. Mean speed and mean absolute yaw rate are reported in $\mathrm{m/s}$ and $\mathrm{rad/s}$, respectively.}
	\begin{tabular}{l||ccccc|cc}
		\noalign{\hrule height 1.0pt}
		\rule{0pt}{2.8ex}\textbf{Configuration} &
		\textbf{SR} (\%)$\uparrow$ &
		\textbf{OSR} (\%)$\uparrow$ &
		\textbf{NE} (m)$\downarrow$ &
		\textbf{SPL} (\%)$\uparrow$ &
		\textbf{CR} (\%)$\downarrow$ &
		\textbf{$\overline{v}$}(m/s)$\uparrow$ &
		\textbf{$\overline{|\omega_z^k|}$}(rad/s)$\downarrow$\\
		\hline\hline
		Single stage: Without distance
		& 79.5 & 80.4 & 12.0 & 64.9 & 17.9 & 1.698 & 0.037\\
		Cruise--Seek: Without distance
		& 76.8 & 83.9 & 16.7 & 59.8 & 12.5 & 1.674 & 0.043\\
		\hline
		Random initial heading
		& 82.1 & 83.0 & 13.6 & 58.1 & 11.6 & 1.528 & 0.059\\
		Without acceleration projection
		& 87.9 & 88.6 & 9.7 & 70.5 & 9.8 & 1.741 & 0.083\\
		\hline
		\rowcolor{gray!15}\textbf{Full configuration (KinoUAV-Agent)}
		& \textbf{89.3} & \textbf{90.8} & \textbf{8.8}
		& \textbf{71.9} & \textbf{6.0} & 1.680 & 0.045\\
		\noalign{\hrule height 1.0pt}
	\end{tabular}
	\label{tab:cruise_seek_ablation}
\end{table*}

\subsubsection{Evaluation of KinoUAV Test Seen}

We evaluate the navigation baselines and different vision-language model (VLM) backbones on KinoUAV Test Seen, as shown in Table~\ref{tab:Test_Seen_comparison}. Random Action fails to complete navigation tasks. CMA achieves only limited navigation success, while TravelUAV improves both SR and obstacle avoidance. AerialVLA achieves the highest SR among the existing methods, but its CR of 28.3\% indicates that reliable obstacle avoidance remains challenging.

The zero-shot experiments show that general-purpose VLMs struggle with this task without fine-tuning. Although increasing the Qwen3-VL model size improves SR, task completion remains limited across all three zero-shot backbones. Fine-tuning substantially improves navigation success and safety. With the low-level controller unchanged, fine-tuning Qwen3-VL-8B increases SR from 3.7\% to 89.3\% and reduces CR from 32.8\% to 6.0\%, owing to its learned waypoint-planning capability. These results demonstrate the importance of adapting the high-level waypoint planner to the KinoUAV task.

All fine-tuned backbones achieve high success rates. The results show that larger models within the same series achieve higher SR and lower CR. Qwen3-VL-8B achieves the highest SR and lowest CR and is therefore used as KinoUAV-Agent. Compared with AerialVLA, it improves SR by 37.0 percentage points and reduces CR by 22.3 percentage points, while achieving a higher mean speed and a lower mean absolute yaw rate. These results show that KinoUAV-Agent improves navigation success and safety, accompanied by faster average motion and reduced heading variation.

\begin{figure*}[t]
	\centering
	\includegraphics[width=\textwidth]{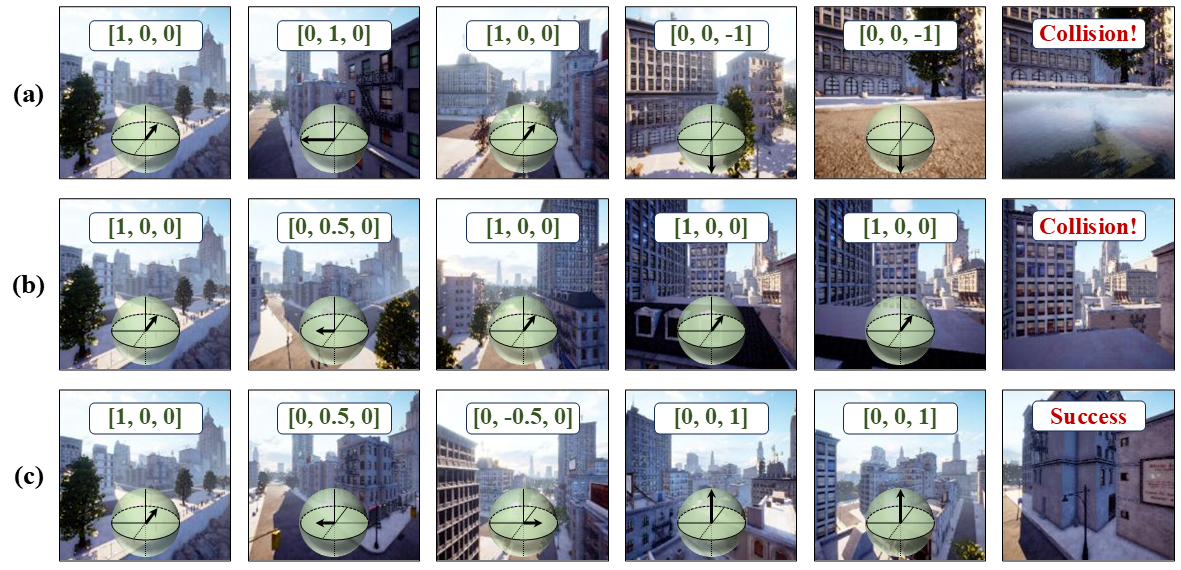}
	\caption{Qualitative comparison on the same navigation episode: (a) without the DFD branch, (b) without DFD supervision, and (c) the full model. The vectors and arrows indicate the selected acceleration actions.}
	\label{fig:dfd_ablation_result}
\end{figure*}

\subsubsection{Evaluation of KinoUAV Test Unseen Object}

We evaluated the model's generalization capability to unseen target objects, and the results are presented in Table~\ref{tab:Test_Unseen_comparison}. KinoUAV-Agent achieved the highest SR of 87.1\%, outperforming the strongest competing method, AerialVLA, by 35.6 percentage points. It also achieved the lowest CR of 6.2\%, whereas TravelUAV—the existing method with the best safety performance—exhibited a collision rate of 17.2\%. These results demonstrate that KinoUAV-Agent can successfully navigate to unseen targets while effectively avoiding collisions.

Among all evaluated methods, KinoUAV-Agent achieved the highest average speed and the smallest heading change, reflecting faster movement and smoother heading adjustments. Compared to the "Test Seen" scenario, there were no significant changes in its success rate, collision rate, or kinematic statistics. This indicates that the navigation strategy learned by the model generalizes effectively to target objects not encountered during training.

\subsubsection{Evaluation of KinoUAV Test Unseen Map}

We further evaluated the model's generalization capability for navigation in unseen scenarios. With the experimental setup remaining unchanged, the results comparing various methods are presented in Table~\ref{tab:Test_Unseen_comparison}; KinoUAV-Agent achieved the highest SR of 64.6\%, outperforming AerialVLA by 25.2 percentage points. Simultaneously, it maintained the lowest CR, demonstrating superior safety compared to existing methods. These results indicate that KinoUAV-Agent retains its advantages in navigation success and safety even in unfamiliar environments. Its average speed and average absolute yaw rate demonstrate consistent motion characteristics across the three test subsets. Overall, the proposed model is capable of generalizing to unseen target objects and environments.

\subsection{Ablation Studies}

We examine how training data, DFD supervision, and navigation strategy affect KinoUAV-Agent. Except for the setting being tested, all variants follow the same training and evaluation settings as the full model. We retrain models when changing their training data, supervision, or prediction targets. Changes to test-time inputs or action quantization do not require retraining. All experiments are conducted on the KinoUAV Test Seen set.

\subsubsection{Analysis of the training dataset}

We first examine the effects of training data composition and size. The Main Subset provides full-navigation trajectories for training both the high-level planner and the low-level controller. Auxiliary Subset provides additional trajectories for training the low-level controller to follow random waypoints. To test the value of these additional trajectories, we fix the high-level planner trained on the full Main Subset. We then retrain the low-level controller using the Main Subset alone or together with Auxiliary Subset. To study the effect of data size, we sample 1\% and 10\% of the training trajectories from each set and compare them with the full-data setting. Both modules are retrained with the sampled data used for their respective tasks.
As shown in Table~\ref{tab:dataset_ablation}, adding Auxiliary Subset increases SR by 18.1 percentage points. Navigation error and collision rate also decrease, while path efficiency improves. The random waypoints require the UAV to follow more varied directions. Training on these trajectories helps the controller track waypoints and avoid obstacles during full navigation.
With only 1\% of the training trajectories, the model rarely completes the task and has the highest collision rate. It also has the lowest mean absolute yaw rate. This suggests that the UAV makes few turns, which may limit its ability to avoid obstacles. Increasing the training data to 10\% improves success rate and reduces collisions. Using the full training data further improves performance.
The model trained on 1\% of the data also performs worse than zero-shot Qwen3VL-8B in Table~\ref{tab:Test_Seen_comparison}. However, the low-level controller in the zero-shot setting is trained on the full training data, whereas both modules in the 1\% setting use limited data. This difference may partly explain the lower performance of the 1\% setting: even when the planner predicts a suitable waypoint, a poorly trained controller may fail to reach it safely.

\begin{table*}[!t]
	\color{black}
	\small
	\centering
	\caption{Analysis of DFD supervision and noise robustness on KinoUAV Test Seen. Noise is added to the supervised DFD variant.}
	\label{tab:dfd_ablation}
	\begin{tabular}{l||ccccc|cc}
		\noalign{\hrule height 1.0pt}
		\rule{0pt}{2.8ex}\textbf{DFD Configuration} &
		\textbf{SR} (\%)$\uparrow$ &
		\textbf{OSR} (\%)$\uparrow$ &
		\textbf{NE} (m)$\downarrow$ &
		\textbf{SPL} (\%)$\uparrow$ &
		\textbf{CR} (\%)$\downarrow$ &
		\textbf{$\overline{v}$}(m/s)$\uparrow$ &
		\textbf{$\overline{|\omega_z^k|}$}(rad/s)$\downarrow$\\
		\hline\hline
		No DFD branch
		& 81.1 & 83.5 & 11.6 & 65.2 & 10.3 & 1.796 & 0.040\\
		Predicted DFD, no supervision
		& 85.7 & 88.4 & 10.3 & 68.9 & 8.5 & 1.791 & 0.044\\
		\rowcolor{gray!15}
		\textbf{Predicted DFD, supervised (KinoUAV-Agent)}
		& \textbf{89.3} & \textbf{90.8} & \textbf{8.8}
		& \textbf{71.9} & \textbf{6.0} & 1.680 & 0.045\\
		\hline
		\quad + noise level 0.1
		& 86.7 & 88.9 & 10.1 & 69.3 & 6.3 & 1.640 & 0.040\\
		\quad + noise level 0.3
		& 85.0 & 86.8 & 12.3 & 68.1 & 8.0 & 1.706 & 0.036\\
		\quad + noise level 0.5
		& 83.3 & 84.8 & 13.6 & 67.8 & 10.7 & 1.741 & 0.040\\
		\noalign{\hrule height 1.0pt}
	\end{tabular}
\end{table*}

\begin{figure*}[t]
	\centering
	\includegraphics[width=\textwidth]{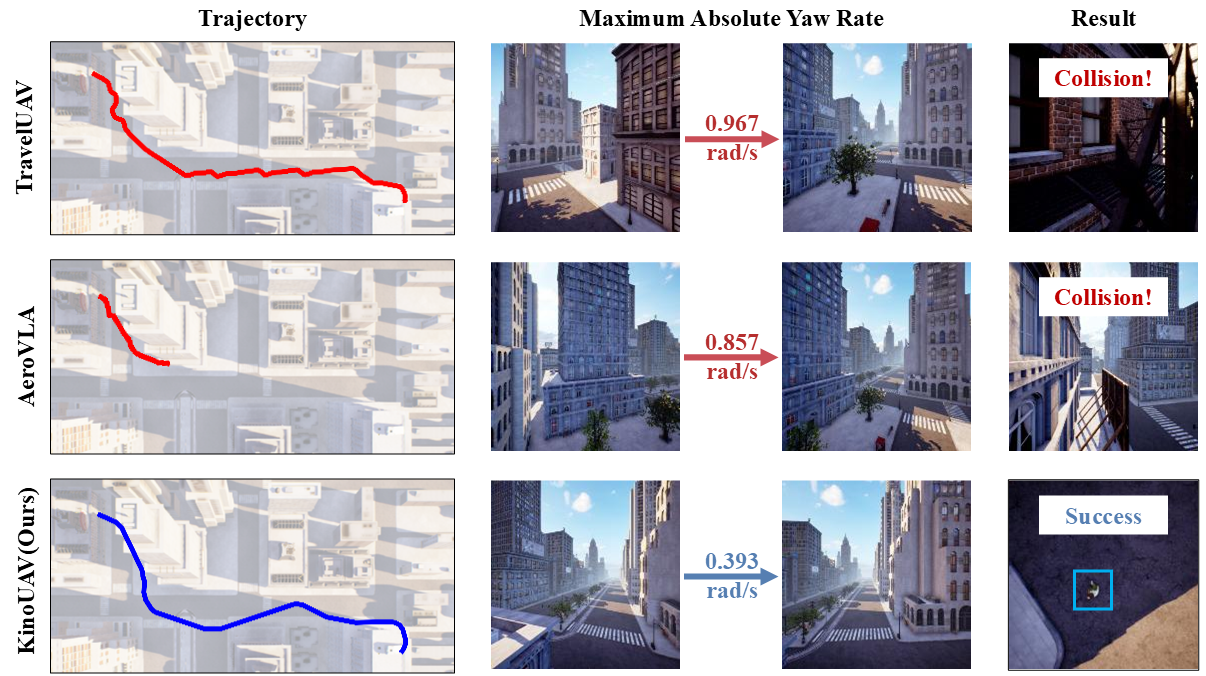}
	\caption{Qualitative comparison of generated trajectories, maximum absolute yaw rates, and navigation outcomes for TravelUAV, AerialVLA, and KinoUAV.}
	\label{fig:exp_result_comparison}
\end{figure*}

\subsubsection{Analysis of the Navigation Strategy}

Table~\ref{tab:cruise_seek_ablation} compares different navigation settings. The full model receives discrete target direction and distance during Cruise and a target description during Seek. Its initial heading is aligned with the next ground-truth waypoint. Before execution, each predicted acceleration is mapped to a predefined action in $\mathcal{A}_{\mathrm{UAV}}$.
We test two changes to navigation guidance. The first removes the Cruise--Seek stages and provides discrete target direction and description throughout navigation, without distance information. The second keeps both stages but removes distance information during Cruise. We retrain the high-level planner for each setting and keep the low-level controller fixed. We also test random initial heading and acceleration execution without quantization. These two changes are applied at test time without retraining.
The single-stage model without distance information has the highest collision rate. Keeping the two stages reduces collisions but also lowers SR. Thus, stage separation alone does not improve task success. Adding distance information improves SR and further reduces collisions. Distance information may help the planner judge how close the UAV is to the target and choose suitable waypoints.
Random initial heading reduces SR by 7.2 percentage points, while increasing CR by 5.6 percentage points. Mean absolute yaw rate increases, while mean speed decreases. The UAV may need to turn more at the start to face the intended route. Initial alignment helps navigation, although the model can still complete tasks with random initial headings.
Removing acceleration quantization gives a similar SR but increases collisions. Mean speed and mean absolute yaw rate also increase, suggesting faster flight and more heading changes. These results show that quantization helps reduce collision risk without a large change in task success. Whether similar output constraints benefit TravelUAV or AerialVLA requires separate experiments.

\subsubsection{Analysis of the DFD}

We first remove the DFD branch and retrain the low-level controller. The acceleration predictor then uses only image features and the navigation state:
\begin{equation}
	\mathbf{a}_{\mathrm{UAV,raw}}^k
	=f'_{\mathrm{action}}\!\left(
	\mathbf{f}_{\mathrm{img}}^k,
	\mathbf{s}_i^k
	\right).
	\label{eq:acceleration_prediction_no_dfd}
\end{equation}
We also train a model that keeps the DFD branch but removes the distance loss:
\begin{equation}
	\mathcal{L}
	=\lambda_{\mathrm{action}}\mathcal{L}_{\mathrm{action}}.
	\label{eq:loss_no_dfd_supervision}
\end{equation}
In this setting, the DFD branch is trained only through the action loss. All other training settings remain unchanged.
To test how DFD errors affect navigation, we add Gaussian noise to the predicted DFDs before passing them to the acceleration predictor:
\begin{equation}
	\mathbf{a}_{\mathrm{UAV,raw}}^k
	=f_{\mathrm{action}}\!\left(
	\mathbf{f}_{\mathrm{img}}^k,
	\mathbf{s}_i^k,
	\mathbf{d}^k+\sigma_n\mathbf{n}^k
	\right),
	\label{eq:acceleration_prediction_noisy_dfd}
\end{equation}
where each component of $\mathbf{n}^k$ is independently sampled from $\mathcal{N}(0,1)$, and $\sigma_n$ controls the noise level. We use the fully trained model and keep its weights fixed.
As shown in Table~\ref{tab:dfd_ablation}, removing the DFD branch lowers SR to 81.1\% and increases CR by 4.3 percentage points compared with the full model. Keeping the branch without distance supervision gives a higher SR and a lower CR than removing it. The model trained with distance supervision performs best. The branch therefore helps action prediction, and directly supervising its distance estimates brings further gains.
As $\sigma_n$ increases from 0 to 0.1, 0.3, and 0.5, SR decreases and NE increases. At $\sigma_n=0.1$, CR remains close to the result without noise, showing that the controller can tolerate small DFD errors. Larger noise leads to poorer navigation, with CR reaching 10.7\% at $\sigma_n=0.5$. These results suggest that the controller uses DFD estimates to select actions and that inaccurate estimates can reduce its ability to avoid obstacles.
Mean speed and mean absolute yaw rate change only slightly across the noise settings. Unlike SR and NE, they do not show a clear trend as noise increases. A lower yaw rate does not always mean better performance: the UAV may turn less but collide more often. In these experiments, the main gains from DFD supervision are a higher success rate and fewer collisions.
Fig.~\ref{fig:dfd_ablation_result} compares the three models on the same test episode. Without the DFD branch, the controller repeatedly selects downward acceleration and eventually hits the ground, as shown in Fig.~\ref{fig:dfd_ablation_result}(a). Without distance supervision, the UAV continues accelerating forward without gaining enough height to clear the roof and collides with it in Fig.~\ref{fig:dfd_ablation_result}(b). The fully supervised model moves sideways and accelerates upward to pass the obstacle, completing the task in Fig.~\ref{fig:dfd_ablation_result}(c). This example shows how DFD supervision helps the controller choose actions that keep the UAV clear of nearby obstacles.

\subsection{Generated Trajectory Analysis}

Fig.~\ref{fig:exp_result_comparison} compares the trajectories of TravelUAV, AerialVLA, and KinoUAV in an urban navigation task. The middle columns show the UAV views near the moment of maximum absolute yaw rate. The rightmost column shows the final result. TravelUAV changes direction repeatedly within a small area. Both TravelUAV and AerialVLA eventually collide with nearby structures, while KinoUAV follows a smoother path along the street and reaches the target.

The maximum absolute yaw rate is $0.393~\mathrm{rad/s}$ for KinoUAV, lower than $0.967~\mathrm{rad/s}$ for TravelUAV and $0.857~\mathrm{rad/s}$ for AerialVLA. This shows that KinoUAV makes less abrupt heading changes in this example. Its high-level planner selects waypoints, and its low-level controller adjusts acceleration to follow them. The resulting trajectory shows how this design allows the UAV to change its motion gradually and pass nearby obstacles safely.

\section{Conclusion}

This paper presents KinoUAV, a benchmark for safe and stable UAV vision-language navigation. KinoUAV includes kinodynamic modeling in an automatic data collection pipeline. It combines global waypoint planning with local kinodynamic A$^*$ search to generate trajectories using acceleration actions. The search accounts for velocity-dependent drag and motion constraints. The benchmark contains 300K trajectories from 16 simulated scenes, divided into the Main Subset and Auxiliary Subset. The former provides full-navigation trajectories, while the latter adds training data for low-level waypoint tracking. Each trajectory includes visual observations recorded with the corresponding motion states and directional free-space distance (DFD) labels.

We also develop KinoUAV-Agent, which uses a high-level planner to predict waypoints and a low-level controller to select acceleration actions. During Cruise, the planner receives the target direction and distance. During Seek, it uses the target description to search for the object. The low-level controller learns to predict DFDs from RGB images and uses these estimates to help select actions. Tests in seen environments, with unseen target objects, and on a held-out map show higher navigation success and fewer collisions than the evaluated baselines. The ablation results show that adding waypoint-tracking data and using more training data improve navigation performance. DFD supervision also helps reduce collisions. Within the Cruise--Seek strategy, distance information improves task success, while a lower yaw rate alone does not guarantee better navigation.

KinoUAV allows researchers to evaluate navigation success together with collision risk and UAV motion. The current benchmark uses simulated scenes and a simplified dynamics model. Future work will include more realistic motion disturbances and moving obstacles. We will also study how the trained models can be transferred to real UAVs.

{\small
\bibliographystyle{IEEEtran}
\bibliography{bibliography}
}

\vspace{-10pt}

\end{document}